\documentclass[sigconf, twocolumn, nonacm]{acmart}
\newcommand\vldbdoi{XX.XX/XXX.XX}

\usepackage{todonotes}[false]
\usepackage{tabularx}
\usepackage{cleveref}
\usepackage{minted}
\usepackage{caption}
\usepackage{rotating}
\usepackage{adjustbox}
\usepackage{pdflscape}
\usepackage[table]{xcolor}
\usepackage{colortbl}  
\usepackage{array,collcell}
\usepackage[ruled,vlined]{algorithm2e}
\usepackage{listings}
 \newcommand{\sspRDFa}{\ensuremath{RDF_{base}}}

\newcommand{\sspRDFc}{\ensuremath{RDF_{llm}}}

\newcommand{\sspJSONa}{\ensuremath{JSON_{base}}}

\newcommand{\sspJSONc}{\ensuremath{JSON_{llm}}}

\newcommand{\sspTEXTa}{\ensuremath{TEXT_{base}}}
\newcommand{\sspTEXTb}{\ensuremath{SSP_{TEXTb}}}
\newcommand{\sspTEXTc}{\ensuremath{TEXT_{llm}}}

\newcommand{\bname}{\texttt{KGI-Bench}}
\newcommand{\tttc}[2]{\texttt{\textcolor{#1}{#2}}}

\newcommand{\imgwidth}{1\linewidth}

\usepackage{pgf} % for pgfmathsetmacro

\newcommand{\colorcell}[1]{%
  \edef\v{\fpeval{max(0,min(1,#1))}}%
  \edef\R{\fpeval{1-\v}}%
  \edef\G{\fpeval{\v}}%
  \edef\doCell{\noexpand\cellcolor[rgb]{\R,\G,0}}%
  \doCell{\v}%
}

\newcolumntype{G}{>{\collectcell\colorcell}c<{\endcollectcell}}

\newcommand{\colorcellf}[1]{%
  \edef\tmp{\fpeval{max(0.6,min(1,#1))}}%
  \edef\v{\fpeval{(\tmp - 0.6)/0.4}}%
  \edef\R{\fpeval{1-\v}}%
  \edef\G{\fpeval{\v}}%
  \edef\doCell{\noexpand\cellcolor[rgb]{\R,\G,0}}%
  \doCell{}#1%
}

\newif\ifshownotes
\shownotestrue      % notes ON
\newcommand{\bluenote}[1]{%
  \ifshownotes
    \textcolor{black}{#1}%
  \else
  \fi
}

\newcolumntype{F}{>{\collectcell\colorcellf}c<{\endcollectcell}}

\begin{document}

% \title{Configuration and Evaluation of Heterogeneous Knowledge Graph Integration Pipelines.}

% \title{Incremental KG Generation: An Empirical Study of Multi‑Source Integration Pipelines}

% \title{\framework: Generation and  Evaluation of Pipelines for Data Integration into Knowledge Graphs}

\title{Evaluation of Pipelines for Data Integration into Knowledge Graphs}

%%
%% The "author" command and its associated commands are used to define the authors and their affiliations.
\author{Marvin Hofer}
\affiliation{%
  \institution{ScaDS.AI Dresden/Leipzig}
  % \streetaddress{P.O. Box 1212}
  \city{Leipzig University}
  \state{Germany}
  \postcode{43017-6221}
}
\email{hofer@informatik.uni-leipzig.de}

\author{Erhard Rahm}
\orcid{0000-0002-2665-1114}
\affiliation{%
  \institution{ScaDS.AI Dresden/Leipzig}
  \city{Leipzig University}
  \country{Germany}
}
\email{rahm@uni-leipzig.de}

\begin{abstract}
Integrating new data into knowledge graphs (KG)  typically involves different tasks that are executed within workflows or pipelines
There are many possible pipelines for a specific integration problem but there is not yet a general approach to evaluate the overall quality and performance of such pipelines to be able to determine the best choices. 
%the lack of systematic evaluation methods for end-to-end knowledge graph (KG) integration pipelines. Existing benchmarks focus on isolated tasks and do not capture interactions across pipeline stages. 
We therefore propose a new benchmark \bname\  to evaluate integration pipelines that ingest different kinds of input data  into an existing KG. We evaluate pipelines by analyzing their output, i.e., the updated KG, with the three complementary quality metrics coverage, correctness and consistency. 
We also provide benchmark datasets (seed KG, overlapping input data of three formats, reference KG as a ground truth) for  the movie domain. To demonstrate the applicability and usefulness of the proposed benchmark, we comparatively evaluate 12 pipelines and analyze their behavior across different input data formats and design choices. 
\end{abstract}

\maketitle

% \pagestyle{\vldbpagestyle}
% \begingroup\small\noindent\raggedright\textbf{PVLDB Reference Format:}\\
% \vldbauthors. \vldbtitle. PVLDB, \vldbvolume(\vldbissue): \vldbpages, \vldbyear.\\
% \href{https://doi.org/\vldbdoi}{doi:\vldbdoi}
% \endgroup
% \begingroup
% \renewcommand\thefootnote{}\footnote{\noindent
% This work is licensed under the Creative Commons BY-NC-ND 4.0 International License. Visit \url{https://creativecommons.org/licenses/by-nc-nd/4.0/} to view a copy of this license. For any use beyond those covered by this license, obtain permission by emailing \href{mailto:info@vldb.org}{info@vldb.org}. Copyright is held by the owner/author(s). Publication rights licensed to the VLDB Endowment. \\
% \raggedright Proceedings of the VLDB Endowment, Vol. \vldbvolume, No. \vldbissue\ %
% ISSN 2150-8097. \\
% \href{https://doi.org/\vldbdoi}{doi:\vldbdoi} \\
% }\addtocounter{footnote}{-1}\endgroup
% %% VLDB block end %%%

% %% do not modify the following VLDB block %%
% %% VLDB block start %%%
% \ifdefempty{\vldbavailabilityurl}{}{
% \vspace{.3cm}
% \begingroup\small\noindent\raggedright\textbf{PVLDB Artifact Availability:}\\
% The source code and datasets have been made available at \url{https://github.com/ScaDS/KGpipe} and \url{https://doi.org/10.5281/zenodo.17246357}.
% \endgroup

% \begin{artifact}
\textbf{Artifact Availability:} \\
All code and datasets are available at: \\
\textbf{Repository:} \url{https://github.com/ScaDS/KGI-Bench} \\  % \todoi{make public}
\textbf{Dataset:} \url{https://doi.org/10.5281/zenodo.17246357} \\
% \end{artifact}

\section{Introduction}

% \todoi{TODO benchmark dataset requirements: incremental, multi-format. Added requirements to \Cref{sec:benchmark}. KGPB is the benchmark/metrics, and KGPB-Mov is the specific instance.}

Knowledge graphs (KGs) have become essential for integrating and representing heterogeneous data in a unified and semantically rich form. Constructing and continuously updating KGs is a  complex task involving solutions for multiple  tasks, such as information extraction, data cleaning, ontology alignment, entity matching and entity fusion. A huge amount of research and development  has led to  powerful tools for these individual tasks that can also benefit from utilizing  Large Language Models (LLMs) ~\cite{DBLP:conf/aitomorrow/MeyerSFRJMDB023,freire2025large}. The task approaches are
typically developed independently and tailored to specific conditions (e.g., data formats)  making it difficult to assemble them into reliable and reusable pipelines to construct and update KGs ~\cite{DBLP:journals/information/HoferOSKR24}. Currently, KG pipelines are mostly handcrafted for specific domains or tasks using ad hoc scripts and configurations. There are a few tools to develop pipelines for constructing and updating KGs but these are mostly not openly available or  restricted to a single application and unable to reuse existing solutions for certain integration tasks ~\cite{DBLP:journals/information/HoferOSKR24}.

A main open challenge lies in evaluating entire KG integration pipelines to identify the quality and problems of such pipelines and to compare the quality and efficiency of different pipelines with each other to help find the best integration approaches.  
This is challenging since pipelines consist of multiple complex tasks that are already difficult to 
evaluate. 
Furthermore the evaluation should be possible for a large spectrum of possible pipelines to enable the integration of diverse data of different formats  and to make use of different implementations for specific tasks including the use of LLMs. Furthermore, it has to be considered that pipelines typically  update an already existing version of a KG in an incremental manner so that input data needs to be matched with each other and the KG to avoid that redundant or duplicate data is introduced in the KG.  

To address this gap, we propose a new benchmark approach called \bname\  (Knowledge Graph Integration Benchmark), to systematically assess end-to-end KG integration pipelines. It focuses on the evaluation of the result of a KG  pipeline, i.e., the newly generated version of a KG  after integrating one or several new input data sources. 
\bname\ uses a set of new evaluation metrics to determine the coverage, correctness and consistency of the updated KG. We also provide a domain-specific instantiation of the benchmark for the movie domain  as a basis for evaluating concrete pipelines with a seed KG, overlapping input data of three data formats (text, JSON , RDF) as well as a reference graph as a ground truth for the quality evaluation. All datasets are made publicly available. 

Our main contributions are as follows:  

\begin{enumerate}
    \item  We propose a general framework for evaluating KG integration pipelines, \bname, including quality metrics for coverage (to what degree is the input data covered in the updated KG), correctness (to what degree is added or changed data in the KG correct)  and consistency of the updated KG. We also consider additional metrics, e.g., regarding pipeline efficiency. Some of the quality metrics are also applicable when no reference graph is available. We further support the weighted combination of  the metrics into a single number to facilitate a ranking among  several pipelines. 
    \item While \bname\  is generic we  provide open benchmark datasets for non-trivial data integration for the movie domain (\bname-Movie). This enables the comparative evaluation of specific pipelines or pipeline tools. The provided benchmark datasets include a seed KG, overlapping input sources in RDF, JSON, and text, as well as a reference KG as a ground truth for the updated KG that includes all input data without duplicates. 
    \item To demonstrate the applicability and usefulness of \bname, we comparatively evaluate 12  integration pipelines (generated by the tool KGpipe \cite{kgpipe25}),  some of which make use of a LLM.  
\end{enumerate}

The remainder of the paper is structured as follows. After reviewing related work, we provide definitions and discuss requirements for end-to-end evaluation of KG integration pipelines. 
We then
introduce the evaluation metrics of \bname. Section \ref{sec:benchmark} presents the benchmark datasets for the movie domain. Subsequently, we describe the evaluated pipelines and their  comparative evaluation using \bname. Finally, we conclude and discuss future directions.

\section{Background and Related Work}

There are several surveys on the approaches for constructing knowledge graphs covering the individual tasks (e.g., data acquisition, ontology design, knowledge extraction,  entity resolution, entity fusion, knowledge completion,) as well as tools to define and execute integration pipelines ~\cite{DBLP:journals/tosem/TamasauskaiteG23,DBLP:journals/information/HoferOSKR24,hogan2021knowledge,weikum2021machine}. We will focus here on previous approaches for evaluating 
data integration for knowledge graphs.
Most of the previous approaches focus on evaluating individual tasks but there also some approaches to evaluate the overall quality and usefulness of knowledge graphs.  
Benchmark datasets play a crucial role in both perspectives by providing common reference points for reproducible comparisons~\cite{zaveri_quality_2015,wang_knowledge_2021,heist_kgreat_2023}.

\textit{
Task-level evaluation} typically relies on ground truth datasets tailored to specific integration tasks. 
For entity resolution, evaluation typically uses gold standards of duplicate records or clusters, with metrics such as  recall, precision and F-measure ~\cite{DBLP:journals/csur/ChristophidesEP21/ERSurvey,DBLP:conf/edbt/0001TTPSSIGPK20/JedAI}. 
Ontology and schema matching tasks  rely on curated reference alignments and are also commonly evaluated using recall, precision, and F-measure  as established in OAEI evaluation campaigns providing different benchmarks ~\cite{euzenat2011ontology,hertlingKnowledgeGraphTrack2020a}.
Recent frameworks such as OntoAligner provide modular support for ontology alignment experiments across multiple benchmark tracks~\cite{giglouOntoAlignerComprehensiveModular2025}.
For schema matching across tabular datasets, evaluation is sometimes ranking-based, using metrics such as Recall@k to measure the retrieval of semantically related columns. 
The Valentine framework supports large-scale experiments for such scenarios~\cite{koutras2021valentine}.
Entity alignment across knowledge graphs uses datasets with known cross-graph correspondences, such as OpenEA, enabling evaluation of  alignment methods using metrics such as Hits@k and MRR~\cite{sun2020benchmarking}. 

For relation extraction, annotated text corpora are used with micro-precision, micro-recall, and micro-F1 metrics, as illustrated by multilingual benchmarks such as RED$^{FM}$~\cite{huguet_cabot_redfm_2023}.
%Benchmark datasets provide the empirical basis for evaluating both individual tasks and entire pipelines. OpenEA is widely used for entity alignment experiments~\cite{sun2020benchmarking}, while the OAEI initiative provides standardized benchmarks for ontology matching. 
Recent benchmarks also support text-to-KG extraction and ontology-driven generation tasks, such as Text2KGBench~\cite{mihindukulasooriya2023text2kgbench} and REDFM~\cite{huguet_cabot_redfm_2023}.
Evaluation datasets and training corpora for extraction, linking, and alignment tasks are often derived from large knowledge bases such as DBpedia and Wikidata  ~\cite{brummerDBpediaAbstractsLargeScale2016}.

%\subsubsection{KG-level Evaluation}
There are different approaches for \textit{evaluating entire KGs} but most of them do not consider the effect of specific integration pipelines. 
Several evaluation dimensions have been proposed to determine the quality of a KG such as correctness, coverage, consistency, timeliness, and usefulness for downstream applications~\cite{zaveri_quality_2015,wang_knowledge_2021,paulheim_knowledge_2016,huaman_steps_2022}. 
Recent work emphasizes that KG quality depends on interactions between data, sources, systems, and usage context~\cite{mohammedFiveFacetsData2025}.
\bluenote{
Recent work has also investigated the evaluation of evolving knowledge graphs and ontologies. Bakker and de Boer propose syntactic and semantic metrics to automatically evaluate changes in knowledge graphs over time, focusing on structural and conceptual quality aspects such as hierarchy structure and semantic consistency between related concepts~\cite{bakker2026dynamic}. Their work mainly targets ontology evolution and schema-level change evaluation, while our work focuses on benchmarking end-to-end KG integration pipelines and evaluating the resulting integrated KG with respect to source coverage, factual correctness, and ontology consistency.
}

Because manual verification of all KG elements (e.g., RDF triples)  is infeasible for large graphs, sampling-based approaches are often used to estimate accuracy (correctness)~\cite{marchesin_efficient_2024}. 
Other methods reduce annotation effort by propagating correctness judgments through logical constraints, as in KGEval~\cite{DBLP:conf/emnlp/OjhaT17/KGEval}. 
In settings without a gold standard, comparative approaches such as ABECTO\cite{keil_abecto_nodate} estimate accuracy and completeness by exploiting overlap between existing knowledge graphs.
\bluenote{Huaman et al.~\cite{Huaman2021Validation} proposed a KG validation approach based on matching entities across weighted external knowledge sources and computing confidence scores from feature similarity comparisons, and FactCheck~\cite{Shami2026LLMvalidation} is a benchmark for LLM-based KG fact validation that evaluates internal model knowledge, retrieval-augmented verification, and multi-model consensus strategies on manually annotated KG triples.}
% Alternative approaches assess the quality of existing overlapping knowledge graphs through inter-graph comparison, for example with ABECTO~\cite{keil_abecto_nodate}, whereas our work evaluates end-to-end integration pipelines that generate such graphs.
% Alternative approaches compare overlapping graphs or external knowledge bases to estimate accuracy and completeness, for example using ABECTO~\cite{keil_abecto_nodate}.\todoi{unclear}

The quality and usefulness of a KG can also be evaluated extrinsically through downstream tasks such as classification or recommendation. 
KGrEaT exemplifies this approach by measuring how different KGs influence task performance~\cite{heist_kgreat_2023}. 
Additional validation approaches include fact-checking frameworks that verify KG assertions using graph structure or external text corpora~\cite{qudus_fact_nodate,DBLP:journals/semweb/0001RS24/ProVe}.

Logical consistency is another important quality dimension. 
Constraint languages such as SHACL~\cite{Cortes2025shacl}, ShEx~\cite{DBLP:conf/www/RabbaniLH22} and RDFUnit~\cite{DBLP:conf/www/KontokostasWAHLCZ14/RDFUnit} allow detection of violations of ontology constraints including domain, range, and cardinality restrictions. 
Sieve combines quality assessment and data fusion by attaching quality scores such as recency or reputation to candidate facts and then using them to select fused values~\cite{DBLP:conf/edbt/MendesMB12/Sieve}.
Other approaches measure inconsistency using logical repair semantics or constraint-based metrics~\cite{dengMeasuringInconsistenciesOntologies2007,bienvenuInconsistencyTolerantQueryingDescription2017,lembo_inconsistency-tolerant_nodate,lukasiewicz_inconsistency-tolerant_2022}. 
Extensions such as PG-Schema bring similar validation capabilities to property graph models~\cite{DBLP:journals/pacmmod/AnglesBD0GHLLMM23/PGSchema}.

%\subsection{Research Gap} 

%Prior work has identified quality dimensions such as completeness, correctness (or accuracy), and consistency for KGs~\cite{paulheim_knowledge_2016, wang_knowledge_2021, Huaman2021Validation}. 
In summary, existing work largely focuses either on conceptual quality dimensions, isolated refinement tasks, or static KG evaluation. Most previous evaluation approaches and benchmarks focus on individual tasks rather than the combined effects of multiple integration stages. Existing KG quality evaluation approaches provide a useful starting point, but lack a focus on incrementally updating a KG through the integration of heterogeneous sources and evaluating the resulting end-to-end integration pipelines. In contrast, \bname\  supports suitable quality dimensions for the end-to-end evaluation of incremental KG integration pipelines. 
%through formally defined entity- and triple-level metrics, alignment-aware comparison, and sequential integration settings. 
To the best of our knowledge, no previous work comparatively benchmarks the quality of different KG integration pipelines in this manner.
% The discussion shows that most previous evaluation approaches and benchmarks focus on individual tasks rather than the combined effects of multiple integration stages. The quality evaluation approaches for entire KGs are a useful starting point, but lack the focus on the effects of incrementally updating a given KG by integrating specific sources of different sources. We are not aware of any previous effort to comparatively benchmark the quality of different integration pipelines. 

\section{Preliminaries}
% (This section introduces our conceptual model and outlines integration challenges across different source types.
%enabling machines to infer new knowledge and support semantic queries. 
We first provide definitions of our notion of a knowledge graph and an integration pipeline. We then discuss requirements for evaluating KG integration pipelines and introduce the metrics of \bname\  to be defined in the next section. 
\subsection{Definitions}
A knowledge graph (KG) is a structured representation of information where entities (such as people, places, or products) are connected through relationships.
An ontology is a formal specification of concepts and the relationships between them within a domain, providing the shared vocabulary and logical rules that underpin knowledge graphs and other semantic systems.
In this work, we assume that KGs are represented in RDF. To limit the problem scope, we study data integration under a fixed KG ontology, leaving dynamic ontology changes for future work.
We refer to individual nodes in the KG as entities, and their assigned entity types correspond to ontology classes. Following common RDF terminology, entities have a unique human-readable name or \textit{label}, e.g., “Titanic (the movie)”. We use the term \textit{property} for any RDF predicate. Properties that link one entity to another are called relations, while properties that link an entity to a literal value are called attributes. Unless the distinction matters, we use property to refer to both relations and attributes. 

Formally, the KG ontology 
$O = (C, P)$ consists of a set $C$  of classes (entity types) and a set $P$ of properties.
Each property $p \in P$ is associated with ontology constraints, including a domain 
$\mathrm{Dom}(p) \in C$, a range $\mathrm{Range}(p) \in C$ for relations, or a datatype 
$\mathrm{Datatype}(p)$ for attributes.
A knowledge graph $KG$ is defined with respect to $O$. We denote its entities by $E(KG)$ and its triples by
\[
T(KG) \subseteq E \times P \times (E \cup L),
\]
where $L$ is the set of literal values. For an entity $e \in E(KG)$, we write $type(e) \subseteq C$ for its assigned types.
% \todoi{bei coverage wurde T(e), und bei consistency type(e) statt c(e) verwendet, type ist am verständlichsten}
We distinguish between \emph{relations}, where the object is an entity, and \emph{attributes}, where the object is a literal.

We also use the term \emph{assertion} for a triple stated in a KG. 
When such an assertion represents domain-level information, such as a film's director or release date, we also refer to it as a \emph{fact}. 
Thus, in the RDF setting used here, facts and assertions are realized as triples.
%while entities and types are represented by RDF resources and type triples.

\textbf{Pipeline Definition}

Given a seed knowledge graph $KG_0$ and one input source $S$, the integration task is to construct an updated graph $KG_1 = \phi(KG_0,S)$. The source may overlap with the seed KG and may use different identifiers, labels, schemas, or representations, requiring tasks such as mapping, alignment, entity resolution, and fusion.

We model a pipeline $\phi$ as a directed acyclic graph (DAG)
\[
\phi = (V_\phi, E_\phi),
\]
where $V_\phi = \{TI_1,\ldots,TI_k\}$ is a set of task implementations and $E_\phi$ denotes directed data-flow dependencies between tasks. Each task implementation $TI_i$ is a function
\[
F_i : D_i^I \rightarrow D_i^O
\]
with well-defined input and output data. A pipeline is valid if, for every task $TI_i$, its required input $D_i^I$ is provided either by the initial inputs $(KG_0,S,O)$ or by outputs of predecessor tasks in $\phi$. This ensures that the pipeline forms a well-defined composition of transformations that produces $KG_1$.

Knowledge graph integration pipelines typically consist of combinations of the following core tasks:

\begin{itemize}
\item \emph{Knowledge Extraction (KE):} transform raw input into structured representations (e.g., triples or candidate entities).
\item \emph{Data Mapping (DM):} map extracted data to the target ontology $O$.
\item \emph{Schema Alignment / Ontology Mapping (SA/OM):} align classes and properties across sources and the target ontology.
\item \emph{Entity Resolution (ER):} identify correspondences between new entities and existing entities in the KG.
\item \emph{Entity Fusion (EF):} merge aligned entities and resolve conflicting information.
\item \emph{Data Cleaning (DC):} detect and correct inconsistencies or errors.
\item \emph{Knowledge Completion (KC):} infer and add missing types, attributes, or relations.
\end{itemize}

These tasks define a common integration workflow, but pipelines may differ in task ordering, implementation choices, and the subset of tasks applied, depending on the source format and integration strategy.

The above definition covers a single source integration step. More complex integration scenarios arise by applying pipelines sequentially to multiple sources. For a sequence of sources $S_1,\ldots,S_n$, this yields
\[
KG_i = \phi_i(KG_{i-1}, S_i), \quad i=1,\ldots,n .
\]
Based on this sequential setting, we distinguish two integration settings. In the \emph{single-source type setting} (SSP), all sources share the same data format (e.g., RDF, JSON, or text), and the same pipeline structure is applied repeatedly across sources. In the \emph{multi-source type setting} (MSP), sources of different formats are integrated sequentially, requiring different pipeline structures or combinations of processing strategies across steps.

Sequential integration in both settings introduces additional challenges such as error propagation, accumulation of inconsistencies, and dependencies of integration order. We later use this setting for incremental evaluation, where quality is analyzed after each integration step.

\subsection{Benchmark Requirements and Metrics Overview}
\label{sec:eval_metrics}

To evaluate knowledge graph (KG) integration pipelines it is not sufficient to assess the isolated performance of individual pipeline tasks but  it is necessary to determine 
the quality of the knowledge graph updated by the complete pipeline. This is because errors introduced at early pipeline stages, such as extraction, mapping, or entity resolution, may propagate, interact, or accumulate during later stages, making task-level metrics insufficient to characterize overall pipeline behavior.

A suitable benchmark for evaluating KG integration pipelines must therefore reflect the characteristics of end-to-end integration and support the computation of suitable metrics to determine the quality of the updated KG w.r.t. the data to be integrated. 
Furthermore, the benchmark should consider that KGs are updated incrementally so that an existing version of a KG is continuously updated to derive newer versions. A benchmark should thus provide an initial seed KG as well as several data sources that are to be integrated by pipelines to create newer versions of the KG. The benchmark should not be  overly complex for the sake of interpretability  but still be realistic requiring integration of heterogeneous data sources that can be dirty and contain overlapping data so that several of tasks mentioned abolve have to be executed in integration pipelines (knowledge  extraction, entity resolution, etc.).

In \bname, we therefore evaluate pipelines at the level of the integrated KG produced after each new source integration and, for the integration of multiple sources,  after the execution of  all pipelines. To evaluate KG quality, we evaluate three complementary quality dimensions: 
%that are widely recognized in the KG quality literature and directly relevant for downstream usage: 
\begin{itemize}
\item \emph{KG coverage}, capturing how completely  source information is represented in the integrated KG; 
\item \emph{KG correctness}, capturing the semantic correctness of added or changed information in the updated KG; and 
\item \emph{KG consistency}, capturing adherence of the updated KG to the formal constraints defined by the target ontology.
\end{itemize}

These dimensions are orthogonal but interacting. A KG may exhibit high coverage but low correctness, for example, by integrating many incorrect facts, or be logically consistent while still missing large parts of the expected information. Consequently, no single metric is sufficient so that meaningful evaluation requires their joint interpretation.

In addition to these core quality dimensions, we consider auxiliary metrics such as structural statistics about the updated KG, pipeline(s) execution runtime, and task-specific metrics such as recall and precision. These metrics support diagnosis, debugging, and performance analysis of pipelines but are not treated as primary indicators of end-to-end KG quality. \bname\  also supports the aggregation of different metrics in a single combined metric to facilitate the comparison of different pipelines. 

We also provide benchmark datasets for the movie domain (\bname-Movie) including a seed KG, source datasets of three formats as well as a reference KG for evaluating coverage and correctness of the generated KG.

\section{\bname\ Evaluation  Metrics}

In \bname,  we assume that the benchmark datasets include a reference KG  as a ground
truth for the integrated KG that includes all input data. We will use the reference graph to determine coverage and correctness (but not consistency) metrics.  Given the difficulty to determine  reference graphs and to underline the general usefulness of the proposed metrics beyond \bname, we will also discuss alternate ways to  determine coverage (using a so-called source-based approach) and correctness (using a labeling-based method).   

We first introduce notations for the definition of the metrics and the alignment between an integrated KG and the reference graph. We then explain the  metrics for the three quality dimensions coverage, correctness and consistency as well as auxiliary metrics.  Finally, we propose a method to aggregate the metrics into a combined metric to allow an easier comparison and ranking of several integration pipelines. 

\subsection{Evaluation Notation and Reference Alignment}

Let $KG_0$ denote the seed knowledge graph, let $S_1, \dots, S_n$ be the source datasets integrated sequentially, and let $KG_n$ be the resulting graph after integrating the first $n$ sources. Furthermore, let $KG_R$ denote the benchmark reference KG. For a dataset or graph $d$, we write $E(d)$ for its entities and $T(d)$ for its RDF triples. We use $KG_R^{(n)} \subseteq KG_R$ to denote the subset of the reference KG that is relevant after the first $n$ integration steps.
%\todoi{wäre es zur Vereinfachung der Notation sinnvoll, die Metriken für einen Integrationsschritt zu erklären und nur zu erwähnen dass diese analog für meherere sukzessive Pipelines anwendbar sind? }
%\bluenote{Depending on the availability of ground truth and the evaluation objective, correctness can be assessed using three complementary strategies: reference-based (alignment), labeling-based, and source-based evaluation.}

\paragraph{Evaluation-restricted entity/triple sets}

To avoid crediting the seed KG for information already present prior
to integration, we restrict evaluation to non-seed content.

We define evaluation-restricted entity and triple sets as:
\[
E(KG_n)_{\mathrm{eval}} = E(KG_n)\setminus E(KG_S),
\]
% \[
% E(KG_R^{(n)})_{\mathrm{eval}} = E(KG_R^{(n)})\setminus E(KG_S),
% \]
\[
T(KG_n)_{\mathrm{eval}} = T(KG_n)\setminus T(KG_S),
\]
% \[
% T(KG_R^{(n)})_{\mathrm{eval}} = T(KG_R^{(n)})\setminus T(KG_S),
% \]
% Where $T_denotes the set of reference triples already represented by the seed KG under the alignment-based triple matching.

and analogously $E(KG_R^{(n)})_{\mathrm{eval}}$ and $T(KG_R^{(n)})_{\mathrm{eval}}$ for the reference KG  by excluding entities and triples
already covered by the seed KG under the alignment.

\paragraph{Reference-based evaluation.}
A direct symbolic comparison between $KG_n$ and $KG_R$ is generally not possible, because equivalent real-world entities may be represented by different identifiers, labels, or literal encodings. This is expected in our benchmark, for example, due to renamed namespaces in RDF sources, generated identifiers for JSON or text sources, and normalization differences in literals. Therefore, evaluation must first establish correspondences between entities in the produced KG and entities in the reference KG.

We model this by an alignment relation
\[
A_n \subseteq E(KG_n) \times E(KG_R),
\]
where $(e,r) \in A_n$ means that entity $e$ in the integrated graph corresponds to entity $r$ in the reference after integration step $n$.

The alignment relation $A_n$ may be obtained using different matching strategies depending on the available signals and the characteristics of the integrated sources. Typical strategies include exact identifier matching, label or literal similarity, provenance tracing across the benchmark construction process, and embedding-based entity similarity. In the context of the benchmark, alignment methods may therefore exploit both structural signals from the integrated KG and provenance information from the dataset generation process. The evaluation framework itself is agnostic to the specific alignment method, requiring only the resulting correspondence relation $A_n$. In our experiments, we derive the alignment relation using embedding-based similarity with sentence-transformer over entity labels and a combination of matching strategies for literal values.

Based on $A_n$, we define the subset of reference entities represented in the integrated graph as
\[
E_R^A(n)=\{r \in E(KG_R) \mid \exists e \in E(KG_n) : (e,r) \in A_n\},
\]
and analogously the subset of integrated entities aligned to the reference as
\[
E_n^A=\{e \in E(KG_n) \mid \exists r \in E(KG_R) : (e,r) \in A_n\}.
\]

\paragraph{Triple Matching Function}
To compare triples, we additionally require literal equivalence. Let $\approx_L$ denote an equivalence relation over literals, e.g., exact equality, normalized string equality, normalized dates, or numerically equivalent values. For triples
\[
t_n=(s_n,p_n,o_n)\in T(KG_n), \qquad
t_R=(s_R,p_R,o_R)\in T(KG_R),
\]
we say that $t_n$ matches $t_R$, written $\mathit{match}(t_n,t_R)$, iff all of the following hold:
\begin{enumerate}
    \item $(s_n,s_R)\in A_n$,
    \item $p_n = p_R$,
    \item either $(o_n,o_R)\in A_n$ for entity-valued objects, or $o_n=o_R$ or $o_n \approx_L o_R$ for literal-valued objects.
\end{enumerate}

Using this definition, the set of matched reference triples is
\[
T_R^A(n)=\{t_R \in T(KG_R) \mid \exists t_n \in T(KG_n) : \mathit{match}(t_n,t_R)\},
\]
and the set of matched produced triples is
\[
T_n^A=\{t_n \in T(KG_n) \mid \exists t_R \in T(KG_R) : \mathit{match}(t_n,t_R)\}.
\]

This yields a two-stage evaluation procedure: first align entities between $KG_n$ and $KG_R$, and then compute coverage and correctness  metrics using the resulting correspondences.

\subsection{KG Coverage}

Coverage measures how completely the source information to be integrated is represented in the integrated KG. 
%Since absolute completeness with respect to the real world cannot be assessed under the Open World Assumption, we measure coverage relative to the benchmark reference data and the known source construction.
%In our benchmark, all seed and source datasets are derived concerning information available in the reference KG with controlled overlap. Hence, for each integration stage, we know which entities and facts are expected to become representable in the result. Coverage therefore quantifies integration completeness rather than real-world completeness.
We distinguish between entity coverage and triple or fact coverage.

\paragraph{Entity coverage.}
% Entity coverage measures how many expected reference entities are represented in the integrated graph:
% \[
% \mathrm{Cov}_E(KG_n)=\frac{|E_R^A(n) \cap E(KG_R^{(n)})|}{|E(KG_R^{(n)})|}.
% \]
Entity coverage measures how many expected reference entities are represented in the integrated graph with the expected semantic type.
Accordingly, a reference entity is counted as covered only if there exists an aligned entity in the integrated graph that is assigned the reference type, since correct typing is important for many downstream tasks. We thus define the set of reference entities covered in the integrated KG as follows:  

\[
\begin{aligned}
E_R^{A,T}(n)=\{\, r \in E(KG_R)\mid {} & \exists e \in E(KG_n): \\
& (e,r)\in A_n \land\ type(r)\subseteq type(e)\,\}.
\end{aligned}
\]
% \[
% E_R^{A,T}(n)=
% \{r\in E(KG_R)\mid
% \exists e\in E(KG_n):
% (e,r)\in A_n
% \land
% type(r)\subseteq type(e)
% \}.
% \]

Note that we do not require that the semantic types of aligned entities are equal but only that the type of the reference entity is included in the set of types of an entity in the integrated KG. This is  because integrated entities may acquire several valid types during integration, for example through type inference or completion.
\bluenote{For instance, an entity extracted as an \texttt{Actor} may additionally obtain the more general type \texttt{Person} through ontology hierarchy inference, while still correctly representing the reference entity. We therefore require that all reference types are preserved while allowing additional compatible types.}

Entity coverage is then defined as
\[
\mathrm{Cov}_E(KG_n)=
\frac{|E_R^{A,T}(n)\cap E(KG_R^{(n)})_{eval}|}
{|E(KG_R^{(n)})_{eval}|}.
\]
considering only the reference entities not already included in the seed KG. 

\paragraph{Fact coverage.}
Fact coverage measures how many expected reference triples are represented in the integrated graph:
\[
\mathrm{Cov}_T(KG_n)=\frac{|T_R^A(n) \cap T(KG_R^{(n)})_{eval}|}{|T(KG_R^{(n)})_{eval}|}.
\]

%Coverage is intentionally evaluated independently of accuracy: an entity or fact counts as covered if it is represented, even if other associated assertions are wrong. This separation is important, since coverage and accuracy describe different failure modes and should therefore be interpreted jointly.

\paragraph{Source-based evaluation}
% Using source information to check correct entailment of knowledge in the KG based on the integrated source.
% The assumption here is that new inforation is correct.

% In addition to reference- and labeling-based evaluation, 
Coverage can also be assessed without reference KG by directly evaluating the integrated KG against the source data to be integrated. In this setting, the source $S_i$ itself serves as the ground truth, and the evaluation verifies whether the information extracted and integrated into the KG is correctly entailed from the source. Formally, let $T(Si) $ denote the set of source facts (after extraction and normalization). $A$ produced triple $t_n \in T(KG_n)$ is considered correct if it can be derived from or matched to a corresponding fact in $T(S_i)$, and for coverage, if for a $T(S_i)$ a matching $T(KG_n)$ is aligned.
%This approach eliminates the need for a reference KG or entity alignment across graphs, and instead evaluates whether the pipeline preserves and correctly represents source information.

This evaluation assumes that the input data is correct and complete with respect to the evaluated facts. While this assumption is reasonable in controlled benchmarks or curated datasets, it may not hold for noisy or real-world sources, where errors can propagate into the KG. Furthermore, this evaluation is inherently local to each source and does not capture cross-source consistency or completeness, making it complementary to reference-based coverage metrics and global consistency checks.
% \todoi{requires de-duplication in the graph itself}

\subsection{KG Correctness}\label{sec:Accu}

The integration of source entities and triples into a KG can lead to many errors, reducing the quality of the resulting KG. Possible errors include unresolved duplicates, false positive entity merges, wrong relation mappings, incorrectly linked literals, or spurious facts introduced during extraction and fusion.
\bluenote{
We use the metrics entity correctness and triple correctness to determine the share of added entities and triples in the integrated KG that are semantically correct relative to the reference KG.
The metrics are precision-oriented, as they quantify the fraction of produced entities/triples that are correct.
}
% We use the metrics entity precision and fact precision to determine the share of added entities and triples of the integrated KG that are correct, which is considered to be the case when we have a match to an entity or triple in the reference KG.  

\paragraph{Entity Correctness}

Entity precision measures for the newly integrated entities the share of correct entities, i.e., entities that are correctly aligned to reference entities and assigned the appropriate entity types.
The produced entities  aligned with the reference KG are defined as 

\[
\begin{aligned}
E_n^{A,T}
= \{\, e \in E(KG_n)\mid {} &
\exists r \in E(KG_R^{(n)}): \\
& (e,r)\in A_n,\;
type(r)\subseteq type(e)\,\}
\end{aligned}
\]
% \[
% E_n^{A,T} = \{e \in E(KG_n) \mid \exists r \in E(KG_R^{(n)}) : (e,r) \in A_n \land T(r) \subseteq T(e) \}
% \]
The entities in this set can still contain duplicates, i.e., entities aligned with the same reference entity.
Duplicates are problematic because they indicate unresolved entity resolution errors. As a result, information about the same real-world object is split across multiple KG nodes, which not only leads to redundancy but can also cause wrong query results and assertions.
To avoid counting all such duplicates as correct, we count the reference entities represented in the integrated KG. 
We define the  aligned reference entities as 

\[
E_{R,n}^{A,T}
=
\{r \in E(KG_R^{(n)})_{eval}
\mid
\exists e \in E_n^{A,T} : (e,r)\in A_n
\}.
\]

Then the entity correctness becomes:

\[
Crct_E(KG_n)=
\frac{|E_{R,n}^{A,T}|}{|E(KG_n)_{eval}|},
\]
The use of reference entities in the nominator ensures that each correctly aligned entity can contribute at most once, while duplicate produced entities increase the count in the denominator, thereby penalizing entity precision. We will also report the duplicate rate as an auxiliary diagnostic metric for explaining unresolved entity resolution errors.

As an auxiliary metric, we also determine  the degree of duplicates by a so-called  
\emph{duplicate rate}. 
A duplicate is present for a reference entity $r$  whenever the set 
\[
\mathrm{Dup}(r)=\{e \in E(KG_n) \mid (e,r)\in A_n\}.
\]
has more than one element. 
%$|\mathrm{Dup}(r)| > 1$.
%Duplicates are problematic because they indicate unresolved entity resolution errors. As a result, information about the same real-world object is split across multiple KG nodes, which leads to redundancy, incomplete query results, and potentially contradictory assertions for what should be one entity.
We measure the duplicate rate as
\[
\mathrm{DupRate}(KG_n)=
\frac{\sum_{r \in E(KG_R)} \max(0, |\mathrm{Dup}(r)|-1)}
{|E(KG_n)|}.
\]

\paragraph{Fact Correctness}
% If you want terminological consistency with the incomplete-reference case, you may want to call this fact precision unless the denominator truly consists of independently labeled produced triples.}

%Because identifiers and literals often differ across sources, exact symbolic comparison is insufficient. 
We again use the alignment relation and triple matching function introduced above. Intuitively, a produced triple is accurate (correct) if it can be matched to a corresponding triple in the reference graph.

As for entity precision, we have to account for duplicate-like triples that match with the same triple of the reference graph. 
%A more restrictive approach counts only one of several triples matching the same reference triple as correct. 
This leads to the following definition of fact precision: 
 
\[
Crct_T(KG_n)=
\frac{|T_{R,n}^{A}|}{|T(KG_n)_{eval}|}.
\]
 
\paragraph{Labeling-based evaluation}

While the above formulation relies on an explicit reference KG and an alignment relation $A_n$ , correctness can alternatively be assessed through direct labeling of entities or triples. In this setting, individual entities or triples in the integrated KG are annotated as correct or incorrect using human experts, LLM-based judgments, or hybrid approaches. This eliminates the need for reference alignment and avoids errors introduced by imperfect matching.
Formally, let $L_n \subseteq T(KG_n)$ denote the subset of labeled triples, together with a labeling function $l : L_n \rightarrow \{0,1\}$ indicating correctness. Correctness can then be estimated as the fraction of correctly labeled triples within $L_n$.

%Compared to reference-based evaluation, labeling-based approaches shift the evaluation from a reference-oriented to an annotation-driven perspective. As a result, coverage metrics with respect to a reference KG are no longer directly applicable, while accuracy metrics become independent of alignment quality and can be applied even when no reference KG exists. 
In practice, labeling is expensive and thus not scalable.  It is therefore often performed on samples only, yielding statistical estimates of correctness rather than measurements. Another limitation is that duplicates are not easily considered so that precision values may become overly high. 

\bluenote{Alternatively, a source-based approach could verify whether a produced KG fact is semantically supported by the source content itself, for example using natural language inference (NLI) methods or LLM-based judges. This is particularly relevant for unstructured or semi-structured sources, where extracted facts may not correspond to exact symbolic matches but can still be supported by the original source text.}

\subsection{KG Consistency}
    % - Positioning relative to prior work
    % - Ontology constraint consistency metrics
    %    O_DT, O_D, O_R, O_RD, O_LT, O_LF
    % - (Duplicates now only auxiliary diagnostic metric if kept)
Consistency measures whether the integrated KG conforms to the structural and semantic constraints defined by the target ontology.
In contrast to coverage and correctness, which compare the produced graph against expected reference content, consistency evaluates whether the resulting graph is internally well-formed with respect to ontology-defined typing, relation usage, and literal constraints.

Ontology consistency has also been analyzed in prior work (e.g. RDFUnit~\cite{DBLP:conf/www/KontokostasWAHLCZ14}, SHACL or SHeX~\cite{DBLP:conf/www/RabbaniLH22}), but it has not yet been investigated as one complementary quality dimension within a benchmark for evaluating end-to-end KG integration pipelines, alongside coverage and correctness.
\bluenote{
In contrast to ontology-evolution approaches such as~\cite{bakker2026dynamic}, which primarily evaluate conceptual or structural changes in existing ontologies, we study consistency as one complementary quality dimension for evaluating the outputs of heterogeneous KG integration pipelines.
}

The consistency metrics defined below capture several common types of ontology violations observed in KG integration pipelines. This set is not intended to be exhaustive. Other ontology constraints, such as cardinality restrictions, functional properties, or key constraints, may also be evaluated depending on the ontology and application context.
% \todoi{reference for the selection, RDFUnit or Ehrlinger Shacl}

Let $\mathrm{type}(e)$ denote the asserted types of entity $e$, and let $\mathrm{Dom}(p)$, $\mathrm{Range}(p)$, and $\mathrm{Datatype}(p)$ denote the ontology-defined domain, range, and datatype of property $p$. We compute the following four violation ratios:

\begin{itemize}
    \item Disjointness violations: entities assigned to mutually disjoint classes
\textit{Example}: an entity typed both as Film and Person.
$$ O_{DT} = \frac{|{e\in E|\exists c1,c2\in T(e):\mathrm{Disjoint}(c1,c2)|}}{|E|} $$

\item Domain and range violations: relations connecting entities whose types do not match the ontology specification 
\textit{Example}: using directedBy to link two Person entities instead of Film → Person.
\[
O_D =
\frac{
\left|
\left\{
(s,p,o) \in R \;\middle|\;
\exists c_s \in T(s): \mathrm{Disjoint}\big(c_s,\mathrm{Dom}(p)\big)
\right\}
\right|
}{|T(KG_n)|}
\]
\[
O_R =
\frac{
\left|
\left\{
(s,p,o) \in R \;\middle|\;
\exists c_o \in T(o): \mathrm{Disjoint}\big(c_o,\mathrm{Range}(p)\big)
\right\}
\right|
}{|T(KG_n)|}
\]

\item Relation direction violations: inverse use of relations relative to ontology definitions
\textit{Example}: asserting Person → Film for a relation defined as Film → Person.
\[
O_{RD} =
\frac{
\left|
\begin{aligned}
\{(s,p,o)\in T(KG_n)\mid\;&
\exists c_s\in \mathrm{T}(s), \exists c_o\in \mathrm{T}(o):\\
&c_s \sqsubseteq \mathrm{Range}(p) \land
c_o \sqsubseteq \mathrm{Dom}(p)
\}
\end{aligned}
\right|
}
{|T(KG_n)|}
\]

\item Datatype and format violations: literal values violating declared datatypes or formats
\textit{Example}: a runtime stored as a free-text string instead of a numeric value, or dates in a different format.
\[
O_{LT} =
\frac{
\left|
\left\{
(s,p,\ell) \in L \;\middle|\;
\neg \mathrm{ValidDatatype}\big(\ell,\mathrm{Datatype}(p)\big)
\right\}
\right|
}{|L|}
\]
\[
O_{LF} =
\frac{
\left|
\left\{
(s,p,\ell) \in L \;\middle|\;
\neg \mathrm{ValidFormat}\big(\ell,\mathrm{Format}(p)\big)
\right\}
\right|
}{|L|}
\]
\end{itemize}

If needed, a similar metric can be defined for format constraints such as year, duration, or currency normalization.

For reporting, these violation ratios can either be presented directly or transformed into compliance scores $C_i = 1 - O_i$, so that higher values uniformly indicate better consistency.

\subsection{Auxiliary Metrics}

In addition to the end-to-end quality metrics discussed above, a variety of auxiliary metrics are commonly reported in KG construction and integration pipelines. These metrics provide useful insights into structural characteristics, resource consumption, and task-specific performance of individual pipeline components. However, they are not suitable as end-to-end quality measures, as they do not directly reflect the semantic correctness, completeness, or logical usability of the resulting KG. In particular, increases in size, density, or computational efficiency do not necessarily correspond to improvements in KG quality and may even mask integration errors. We therefore exclude these metrics from our end-to-end quality assessment and report them only as diagnostic and explanatory indicators for pipeline analysis, debugging, and scalability evaluation. 
%but must be interpreted independently from semantic KG quality.
\paragraph{Statistical Metrics}

These metrics capture the structural properties of the generated KG, giving a baseline view of its size and complexity (more classes or properties).  
% All values are normalized with respect to the seed KG (0.0) and the reference KG (1.0) where applicable.  

We measure the structural characteristics of the generated KG:

\begin{itemize}
    \item $\mathbf{S_{FC}}$ \textbf{Fact Count}: Number of distinct triples (facts).
    \item $\mathbf{S_{EC}}$ \textbf{Entity Count}: Number of distinct entities (nodes).
    \item $\mathbf{S_{RC}}$ \textbf{Relation Count}: Number of distinct relation names.
    \item $\mathbf{S_{TC}}$ \textbf{Type Count}: Number of distinct entity classes (types).
    \item $\mathbf{S_{UT}}$ \textbf{Count of untyped entities} that remain without a class, e.g., because none of their extracted or mapped properties match the ontology. %When only a label is retained and no valid relations exist, 
    In such cases, type inference may not be able to assign a class leading to 
%    , and fusion produces an 
    isolated label-only entities.
%    \todoi{muss das Fehlen der Klasse wirklich das Fehlen aller properties/relations bedeuten? Ja, muss es, da die Ontology für alle Properties (außer label oder type) die Domain (falls subject) und Range (falls object) explizit auf eine der drei Klassen definiert.}
    \item $\mathbf{S_D}$ \textbf{Graph Density}: Ratio of existing relations to the maximum possible number of relations between entities, indicating overall connectivity.  Identifying most relations results in a KG with high density while a KG with mostly isolated entities has low density.
\end{itemize}
% To normalize these values and for calculating an average score we choose to build an normalized interval between the estimated reference values (or higher) as 1.0 and the values of the seed KG.

\paragraph{Resource Metrics}

These metrics capture the computational cost and efficiency of generating the KG (Quality-of-service).  

\begin{itemize}
    \item $\mathbf{Q_D}$ \textbf{Duration}: Total execution time of the pipeline in seconds.
    \item $\mathbf{Q_M}$ \textbf{Max Memory}: Peak memory consumption during runtime.
    \item $\mathbf{Q_C}$ \textbf{Additional Costs}: Non-computational overhead such as API usage, cloud hosting, or power consumption.
\end{itemize}

% TODO: Table or report format for semantic violations by pipeline

\paragraph{Task Metrics}

For each task in a pipeline there can be a set of specific evaluation metrics. In addition to the duplicate rate already introduced in subsection \ref{sec:Accu}, \bname\  considers:   

\begin{itemize}

    \item $\mathbf{R_{EM}}$ \textbf{Entity Matching}: Precision and recall of expected vs.\ actual entity correspondences in entity resolution tasks ~\cite{DBLP:journals/csur/ChristophidesEP21/ERSurvey}.
    \item $\mathbf{R_{OM}}$ \textbf{Ontology Matching}: Precision and recall of expected vs.\ actual relation/property correspondences in ontology matching tasks.
    \item $\mathbf{R_{EL}}$ \textbf{Entity Linking}: Coverage of links between entities in the KG and external reference identifiers (e.g., ensuring each root entity is linked to at least one external reference).
    \item $\mathbf{R_{RL}}$ \textbf{Relation Linking}: Correctness of links between input features (e.g., JSON keys) and ontology relations, where gold mappings are available. 
\end{itemize}

\subsection{Aggregation and Ranking Method}
% \todoi{remove/change?, harmonical mean vs average?}
    % - Group aggregation
    % - Weighted ranking

To better compare and rank different pipelines, it is desirable to aggregate the different metrics in fewer scores or even a single score per pipeline. 
A simple way is to determine the F1-scores for entities and triples  as the harmonic means of  their coverage (recall) and precision metrics. 
To combine more metrics in a single score, \bname\ supports a two-step aggregation of metrics.
%grouping previously described performance and quality metrics into account. 
We first normalize and aggregate selected metrics per group (e.g. quality dimension)  and then determine a weighted average of the group metrics to obtain the overall aggregated score of a pipeline. 

Each group metric is the average of the  normalized metrics $Mn{_i}$ (mapped to interval [0,1], higher is better) of the selected individual metrics $M{_i}$ per group: 
\[
GM_i = \frac{Mn_1+\dots+Mn_k}{k}.
\]

The total aggregated evaluation score $M_{\text{total}}(p)$  of a pipeline $p$ is the weighted average of the  group metrics:  
\[
M_{\text{total}}(p) = \sum_{i=1}^j w_i \cdot GM_i(p), \quad \sum_{i=1}^j w_i = 1,
\]

When focusing on evaluating the three quality dimensions KG coverage, correctness,  and consistency, we obtain:
\[
M_{\text{total}}= \alpha \cdot GM_{\text{cov}} + \beta \cdot GM_{\text{acc}} + \gamma \cdot GM_{\text{cons}} %$+ \delta \cdot GM_{\text{4}}
\]
where $\alpha, \beta, \gamma, \delta$ are weights that can be adjusted depending on the user’s goals (e.g., coverage vs consistency).
We will use this approach in the evaluation  in~\Cref{sec:ranking}.

\section{\bname-Movie}
\label{sec:benchmark}

The \bname\   evaluation metrics  can be applied to integration tasks for  knowledge graphs of different domains and application purposes. For specific and comparable evaluation results, we need however a well-defined  KG and source datasets to be integrated of a certain domain. In this section we therefore outline an  instantiation of \bname\  for the movie domain (\textbf{\bname-Movie}) and how we created the datasets. Benchmark datasets for other domains can be constructed in an analogous way. The next section outlines example integration pipelines that we  use in our evaluation with  \bname-Movie (Section \ref{sec:results}). 

The \bname-Movie\ datasets cover films and their related persons and companies. The domain provides a manageable yet realistic setting with heterogeneous data formats, partial overlap between sources, and schema-level heterogeneity.
\bluenote{
While existing benchmarks such as Text2KGBench~\cite{mihindukulasooriya2023text2kgbench} evaluate isolated text-to-KG extraction tasks, KGI-Bench-Movie targets incremental end-to-end KG integration across heterogeneous RDF, JSON, and text sources, including challenges such as entity resolution, fusion, and ontology consistency.
}
While the benchmark is domain-specific, it represents a general class of incremental KG integration scenarios across heterogeneous sources of three formats (RDF, JSON, and text). We first describe the domain and KG ontology before providing details about the datasets (reference KG, seed KG, sources with overlapping entities) and their generation. 

\paragraph{Domain and Ontology}

Our benchmark for the movie domain includes entities such as films, actors, directors, and production companies. We manually curated a target ontology for this domain  comprising:

\begin{itemize}
    \item Three core classes: \texttt{Film}, \texttt{Person}, and \texttt{Company}, serving as the backbone of the domain model.  
    \item 25 properties (object relations or datatype properties, and counting the RDF \textit{label} and \textit{type} property), defined using OWL, RDF Schema, and SKOS, enabling consistent links across heterogeneous sources.  
    \item Annotations such as \texttt{rdfs:label} and \texttt{skos:altLabel} (to support lexical variation) and \texttt{owl:equivalentClass} or \\ \texttt{owl:equivalentProperty}) that facilitate schema alignment.  
    \item Integrity axioms, including \texttt{owl:disjointWith} (to catch invalid class overlaps, e.g., a \texttt{Film} typed as \texttt{Person}) and \texttt{owl:maxCardinality} (to follow rules, e.g., one \texttt{runtime} per film).  
\end{itemize}

\begin{figure}
    \centering
    \includegraphics[width=\imgwidth]{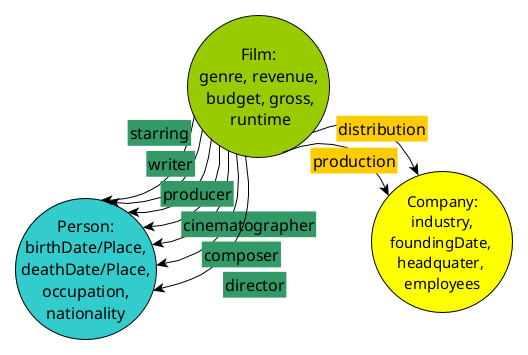}
    \caption{Ontology/Schema graph of classes film, person, company with their properties (relations and attributes).}
    \label{fig:ontology}
\end{figure}

A visualization of this graph schema (ontology) is shown in \Cref{fig:ontology}.  

%\paragraph{Essential}
A well-defined ontology is essential for the integration process because it provides the common target schema against which heterogeneous sources can be aligned. In the linking step, ontology classes and relations guide entity and schema matching, ensuring that different source fields map to the correct concepts. For cleaning, constraints and disjointness axioms allow the detection of errors such as a film mistakenly typed as a person or multiple runtimes assigned to the same movie. Finally, during completion, the ontology defines where additional values can be added consistently, for example, enriching a person entity with missing \texttt{birthPlace} or linking a film to its \texttt{director}. In short, the ontology acts as the semantic backbone that enables heterogeneous data to be reconciled into a coherent knowledge graph.

%\paragraph{Complexity}
Although the ontology provides a clear target schema, integration remains challenging due to ambiguity and inconsistency in the sources. For example, some JSON infoboxes use the field name "date" without specifying whether it denotes a person’s birth or death date, which are distinguished in the ontology. 
%between \texttt{birthDate} and \texttt{deathDate}. 
% Similarly, the property \texttt{producer} may overlap with the field \texttt{production}, which in some cases refers to different entity types.
Ambiguities also arise among role-related person or company properties for a film, e.g, \texttt{producer} as person vs \texttt{production} as company.
% properties. The ontology distinguishes six different person relations and two company relations  \texttt{production} and \texttt{distribution}.  
Properties may also use different data types. Runtime may appear in minutes, hours, or as free-text strings (“2h 15m”), and financial attributes such as \texttt{budget}, \texttt{gross}, and \texttt{revenue} are reported with inconsistent currencies and not always distinguished across sources.

% These examples illustrate how semantic heterogeneity and ambiguity create real obstacles in mapping source data into the target ontology, motivating the need for robust pipelines and validation strategies.

\textit{Datasets} Our benchmark is built from a set of interrelated data artifacts that provide the basis for constructing and evaluating pipelines. Starting from a reference graph, we derive seed and source datasets in multiple formats (RDF, JSON, and text), each designed to expose specific integration challenges. These artifacts vary in coverage, structure, and representation, thereby reflecting the heterogeneity typical of real-world scenarios. 
% A summary of their scale and composition is provided in \Cref{tab:test_dataset_stats}.

\textit{Reference KG.} To establish a basis for evaluation, we first derive a reference by recursively collecting data for the film entities from DBpedia, and their related persons and companies, whenever such links are provided by the eight ontology relations.
% This results in a comprehensive KG that serves as the gold standard for downstream evaluation.

We restrict our benchmark to 10,000 films rather than the full set of ~150,000 available in DBpedia (the final size is a bit lower due to cleaning and generating overlaps). This size already yields a knowledge graph of substantial complexity once related persons and companies are included, while keeping experiments computationally feasible and reproducible. It offers sufficient semantic variety to capture typical integration challenges without introducing unnecessary volume. For development and reduced resource requirements, we provide two smaller benchmark versions   with only a total of 100 and 1,000 film entities. 

From the reference graph, we generate multiple entity splits to emulate realistic multi-source settings. Specifically, we partition the set of all films into four subsets (seed and three sources) of 25\%  of the entities each. To introduce controlled redundancy, every pair of subsets (seed, sources) shares around 5\% of their film entities (30\% including persons and companies). This controlled overlap reflects the partial coverage commonly observed in real-world sources, where new sources usually cover existing entities in the KG.

\textit{Seed/Source type RDF.} The seed KG is defined as the first split of the reference graph. To generate the corresponding source RDFs, we create shaded versions of the remaining splits by renaming entity identifier namespaces. This ensures that the same real-world entities are represented with different IRIs across splits, thereby requiring entity resolution during integration.

\begin{lstlisting}[
    % frame=single,
    basicstyle=\small %\tiny or \small or \footnotesize etc.
]
# Source RDF-KG
<@\tttc{cyan}{rdf:Titanic}@> a <@\tttc{cyan}{rdf:Film}@>; <@\tttc{black}{rdfs:label}@> <@\tttc{cyan}{"Titanic"}@>;
  <@\tttc{cyan}{rdf:actor rdf:DiCaprio}@> .
<@\tttc{cyan}{rdf:Diamonds}@> a <@\tttc{cyan}{rdf:Film}@>; <@\tttc{black}{rdfs:label}@> <@\tttc{cyan}{"Diamonds"}@>;
  <@\tttc{cyan}{rdf:actor rdf:Douglas}@> . 
\end{lstlisting}

\textit{Source type JSON.} For each film entity, we generate nested JSON records derived from 
% Wikipedia Infobox properties (raw key-value pairs).\todo{no longer true} 
one subgraph per film, including referenced person and company information.
Each record contains key–value pairs describing the film, such as title, actors, genre, or production company. 
%Unlike the RDF representation, the JSON data is intentionally flatter: attributes referring to persons or companies are included only by name rather than as nested objects. This makes 
JSON is a suitable source format for testing pipelines, since it is widely used on the Web, comes with less structural richness than RDF, and requires additional mapping to be aligned with the ontology.

% JSON 
\begin{lstlisting}[
    basicstyle=\small %\tiny or \small or \footnotesize etc.
]
# JSON Document
<@\tttc{orange}{\{ "title" : "Inception",}@>
  <@\tttc{orange}{"runtime" : "8880.00",}@>
  <@\tttc{orange}{"starring" : [}@>
    <@\tttc{orange}{\{ "name" : "Leonardo DiCaprio",}@>
      <@\tttc{orange}{"birthYear" : ... \} ... ] ... \}}@>
\end{lstlisting}
% # Mapped RDF-KG
% <@\tttc{orange}{json:Inception}@> a <@\tttc{orange}{json:Film}@>; <@\tttc{black}{rdfs:label}@> <@\tttc{orange}{"Titanic"}@>; 
%   <@\tttc{orange}{json:actor json:DiCaprio}@> . 
%   <@\tttc{orange}{json:runtime "8880.00"}@>^^<@\tttc{black}{xsd:double}@> .
  
\textit{Source type Text.} As a complementary unstructured source, we include DBpedia abstracts for all film entities. These textual descriptions provide narrative information mentioning films, people, and companies without explicit schema structure. Text is an important source type because it reflects the reality that much of the Web’s knowledge is available only in natural language and requires information extraction to be transformed into a knowledge graph.

\begin{lstlisting}[
    basicstyle=\small %\tiny or \small or \footnotesize etc.
]
# Text Document
<@\tttc{teal}{Titanic}@> <@\tttc{black}{is a}@> <@\tttc{green}{1997}@> <@\tttc{black}{American}@> <@\tttc{green}{epic historical romance}@> <@\tttc{teal}{film}@>
<@\tttc{olive}{written}@> <@\tttc{black}{and}@> <@\tttc{olive}{directed}@> <@\tttc{black}{by}@> <@\tttc{teal}{James Cameron}@>.
<@\tttc{black}{Incorporating both historical and fictional aspects, it is}@>
<@\tttc{black}{based on accounts of the sinking of RMS Titanic in 1912.}@>
<@\tttc{teal}{Leonardo DiCaprio}@> <@\tttc{black}{and}@> <@\tttc{teal}{Kate Winslet}@> <@\tttc{olive}{star}@> <@\tttc{black}{as}@> <@\tttc{black}{members of...}@>
\end{lstlisting}
% # Extracted RDF-KG
% <@\texttt{
% \textcolor{teal}{txt:Titanic} a \textcolor{teal}{txt:Film}; rdfs:label "Titanic"; \\
% \textcolor{green}{txt:release} "1997"xsd:gDate; \\
% \textcolor{green}{txt:genre} "epic", "historical romance"; \\
% \textcolor{olive}{txt:actor} \textcolor{teal}{txt:DiCaprio}, \textcolor{teal}{txt:Winslet}; \\
% \textcolor{olive}{txt:writer} \textcolor{teal}{txt:Cameron}; \textcolor{olive}{txt:director} \textcolor{teal}{txt:Cameron} .
% }@>

\textit{Complementarity of Sources.} Together, RDF, JSON, and text capture the three principal degrees of structure encountered in real-world data: highly structured (RDF), semi-structured (JSON), and unstructured (text). This diversity ensures that the benchmark exercises pipelines across the full spectrum of integration challenges, from ontology alignment and schema mapping to entity resolution and information extraction.  

\textit{Supplementary Data.} In addition to the reference, seed, and source datasets, we provide supplementary data that supports evaluation and error analysis. First, we include metadata about the expected matches, i.e., the overlapping entities introduced during the dataset splitting. These records specify the entity type (e.g., \texttt{Film}) together with explicit links of the form \texttt{id1 = id2}, which serve as ground truth for entity resolution tasks. Second, we provide curated lists of verified source entities. These lists indicate which entities should appear in the integrated KG, enabling evaluators to check for missing or erroneously introduced entities. Together, these supplementary resources facilitate a more fine-grained assessment of pipeline performance beyond the structural and semantic properties of the resulting graphs. 

We end the section with two additional examples: (i) a targeted Seed KG and (ii) the resulting integrated RDF KG obtained after incorporating the three previously shown source snippets into the seed. Color highlights indicate the origin of each entity (i.e., from the seed or from a specific source). In this example, no conflicting attribute values occur during integration.
\begin{lstlisting}[
    basicstyle=\small, %\tiny or \small or \footnotesize etc.
]
# Seed RDF-KG
<@\tttc{black}{kg:Titanic a kg:Film}@>; <@\tttc{black}{rdfs:label "Titanic"}@>;
  <@\tttc{black}{kg:runtime "11700.00"}@>^^<@\tttc{black}{xsd:double}@> .
  <@\tttc{black}{kg:actor}@> <@\tttc{black}{kg:DiCaprio}@> .
<@\tttc{black}{kg:Matrix}@> a <@\tttc{black}{kg:Film}@>; <@\tttc{black}{rdfs:label "The Matrix"}@>;
  <@\tttc{black}{kg:release "1999"}@>^^xsd:gDate;
  <@\tttc{black}{kg:genre "science fiction", "action film"}@>;
  <@\tttc{black}{kg:writer kg:Wachowskis; kg:director:Wachowskis}@>;
  <@\tttc{black}{kg:actor kg:Reeves}@> . 

# Integrated Result RDF-KG
<@\tttc{black}{kg:Titanic a kg:Film}@>; <@\tttc{black}{rdfs:label "Titanic"}@>;
  <@\tttc{black}{kg:runtime "11700.00"}@>^^<@\tttc{black}{xsd:double}@> .
  <@\tttc{black}{kg:release}@> <@\tttc{green}{"1997"}@>^^<@\tttc{black}{xsd:gDate}@>;
  <@\tttc{black}{kg:genre}@> <@\tttc{green}{"epic", "historical romance"}@>; 
  <@\tttc{black}{kg:writer}@> <@\tttc{teal}{txt:Cameron}@>; <@\tttc{black}{kg:director}@> <@\tttc{teal}{txt:Cameron}@> .
  <@\tttc{black}{kt:actor kg:DiCaprio}@>, <@\tttc{teal}{txt:Winslet}@> .
<@\tttc{black}{kg:Matrix}@> a <@\tttc{black}{kg:Film}@>; <@\tttc{black}{rdfs:label "The Matrix"}@>;
  ... <@\tttc{black}{kg:actor kg:Reeves}@> . 
<@\tttc{orange}{ex:Inception}@> a <@\tttc{black}{kg:Film}@>; <@\tttc{black}{rdfs:label}@> <@\tttc{orange}{"Inception"}@>; 
  <@\tttc{black}{kg:runtime}@> <@\tttc{orange}{"8800"}@>^^<@\tttc{black}{xsd:double}@> .
  <@\tttc{black}{kg:actor kg:DiCaprio}@> .
<@\tttc{cyan}{ex:Diamonds}@> a <@\tttc{black}{kg:Film}@>; <@\tttc{black}{rdfs:label}@> <@\tttc{cyan}{"Diamonds"}@>;
  <@\tttc{black}{kg:actor}@> <@\tttc{cyan}{ex:Douglas}@> . 
\end{lstlisting}

%For development and reduced resource requirements. We provide two smaller benchmark versions   with only a total of 100 and 1,000 film entities.  

% https://app.diagrams.net/#G1fliRaNP5pcAR0oU1b06Q5vtk7DtJUvrh#%7B%22pageId%22%3A%22y04jnIRG2x4srCmvvEEI%22%7D
\begin{figure}
    \centering
    \includegraphics[width=\imgwidth]{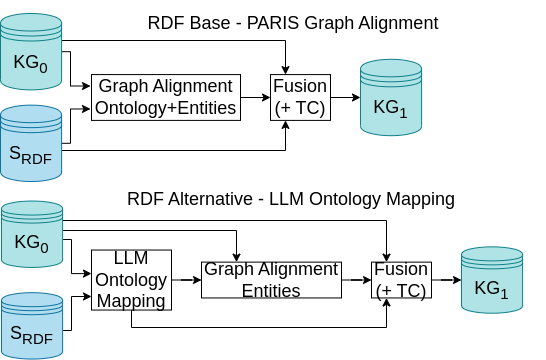}
    \caption{RDF single-source pipeline layouts used in the evaluation. TC=Type Completion.}
    \label{fig:ssp_rdf}
\end{figure}

\begin{figure}
    \centering
    \includegraphics[width=\imgwidth]{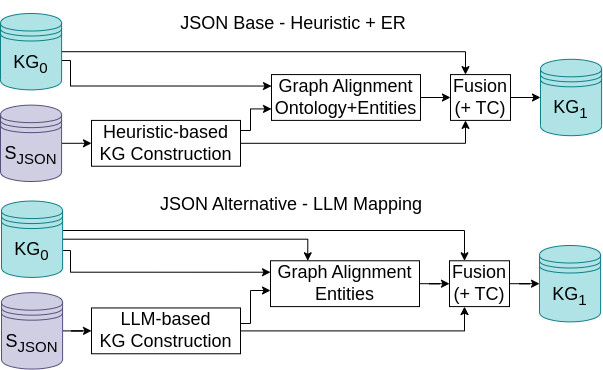}
    \caption{JSON single-source pipeline layouts used in the evaluation. TC=Type Completion}
    \label{fig:ssp_json}
\end{figure}

\begin{figure}
    \centering
    \includegraphics[width=\imgwidth]{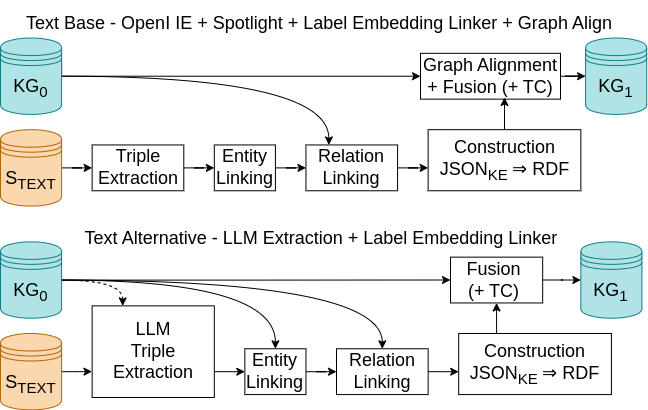}
    \caption{Text single-source pipeline layouts used in the evaluation. TC=Type Completion}
    \label{fig:ssp_text}
    % \todoi{the variants are different in two tasks instead of only one as for rdf and json}
\end{figure}

\section{Evaluated Pipelines}
\label{sec:pipelines}
%\todoi{naming of pipeline variants: $RDF_{base}$, $RDF_{llm}$}

To apply the proposed benchmark, we define a set of knowledge graph integration pipelines operating on the source formats supported by \bname-Movie: RDF, JSON, and text. The pipelines  are not intended to be optimal solutions, but serve as representative integration strategies with different design choices based on existing task implementations or LLM functionality.
For each source format, we define two pipeline variants: (i) a baseline pipeline using standard techniques, and (ii) a variant in which selected tasks are replaced by LLM-based implementations. All pipelines have been specified and executed with the open-source tool KGpipe \cite{kgpipe25}.
%This allows us to analyze the effect of different implementation strategies within a comparable pipeline structure.

\Cref{fig:ssp_rdf,fig:ssp_json,fig:ssp_text} illustrate the corresponding six pipeline layouts for RDF, JSON, and text sources, respectively. 
%While the overall workflow is similar across formats, the role and difficulty of individual tasks differ substantially.
All pipelines follow the general workflow introduced in Section~3.1, comprising tasks such as knowledge extraction, data mapping, schema/ontology alignment, entity resolution, and fusion to address common integration challenges such as schema heterogeneity and ambiguity in extraction and matching. Depending on the source format and implementation, some of these steps may be simplified, combined, or omitted.
Pipelines combine established tools with lightweight custom transformations that ensure interoperability between tasks by converting outputs into the required input formats for subsequent tasks. 
%These transformations enable the composition of heterogeneous components for matching and fusion. 
The LLM-based variants replace selected steps such as extraction, mapping, or ontology matching with generative approaches. In the following we discuss the pipelines for the three input formats.
% To ensure interoperability and comparability, we use two standardized intermediate representations: $JSON_{ER}$ for entity correspondences and $JSON_{KE}$ for extracted knowledge.

%Across all source types, pipelines must address common integration challenges such as schema heterogeneity, ambiguity in extraction and matching, and error propagation across stages.
% Therefore, all pipelines include standard post-processing steps such as normalization, duplicate handling, and type completion, ensuring that differences in evaluation results primarily reflect pipeline design choices rather than missing functionality.

\subsection{Structured Sources (RDF)}  

For structured sources, integration primarily focuses on schema alignment and entity resolution, as data is already available in a graph-like representation. A typical pipeline is:

\emph{(i) Data Mapping (to RDF for relational input) $\rightarrow$ (ii) Schema Alignment $\rightarrow$ (iii) Entity Resolution $\rightarrow$ (iv) Entity Fusion $\rightarrow$ (v) Cleaning $\rightarrow$ (vi) KG completion (optional).}

%\paragraph{RDF Pipelines}
We evaluate two RDF pipeline layouts, each with one configuration (see~\Cref{fig:ssp_rdf}):

\begin{itemize}
    \item \sspRDFa~ applies the established  graph alignment method PARIS~\cite{DBLP:journals/pvldb/SuchanekAS11/Paris} (executed in a Docker container) to produce matches between entities in the seed and source RDF graphs. A fusion algorithm with a first-value preference is then applied for matching entities: entity identifiers are resolved from source to seed, and for fusable relations, the first available value (typically from the source) is selected. Afterwards, type information on entities is inferred based on their current properties corresponding domain/range specs.
    % \item \sspRDFb~ first transforms both RDF graphs into tabular CSV representations using a custom Python function. It then applies record linkage and schema matching to detect similar entities and relation names, using the clean–clean implementation of JedAi~\cite{DBLP:conf/edbt/0001TTPSSIGPK20/JedAI} and Valentine~\cite{DBLP:conf/icde/KoutrasSIPBFLBK21/Valentine} for matching. Finally, the fusion algorithm is applied as in \sspRDFa.
    \item \sspRDFc~ follows the same structure as \sspRDFa, but determines relation alignment (ontology matching) with the help of a Large Language Model (LLM). The LLM is used to match and map relations between the source and seed RDF graphs, based on sampled triples from each KG. PARIS is then applied for entity matching, followed by the same first-value fusion strategy.
\end{itemize}

\subsection{Semi-Structured Sources (JSON)}

%For semi-structured sources, the main challenge is interpreting implicit schema and mapping heterogeneous structures to the target ontology.
Semi-structured data introduces complexity through implicit or evolving schema, nested records, and heterogeneous field names. 
Converting JSON, XML, or CSV into graph-compatible form requires therefore careful parsing and normalization, often combined with type inference to recover implicit semantics.
A central challenge here is schema interpretation and mapping ~\cite{van_assche_declarative_2023,bizer2004d2rq}, as the source structure may only partially reflects the target ontology. 
%Concretely, this challenge is addressed through both declarative mapping ~\cite{van_assche_declarative_2023,bizer2004d2rq}approaches~\cite{van_assche_declarative_2023} and heuristic or learning-based techniques~\cite{bizer2004d2rq} for schema interpretation and mapping of JSON, XML, and CSV sources.
%Once this transformation is complete, subsequent resolution and fusion can build directly on techniques from the structured case.

 A typical pipeline is:

\emph{(i) Parsing and Normalization $\rightarrow$ (ii) Data Mapping (to RDF) $\rightarrow$ (iii) Type Inference / Schema Alignment $\rightarrow$ (iv) Entity Resolution $\rightarrow$ (v) Fusion $\rightarrow$ (vi) Cleaning.}

%\paragraph{JSON Pipelines}
For the JSON format, we also have two layouts with a configuration each (see~\Cref{fig:ssp_json}). 

\begin{itemize}
    \item \sspJSONa~ maps JSON data into a generic RDF graph, each key as a generic predicate URI, and types are constructed from the current path keys.
    %using a custom heuristic. 
    The implementation attempts to identify \texttt{rdfs:label} values from key–value pairs in JSON objects and distinguishes between shallow objects (treated as entities with labels) and literals (e.g., dates, numbers) using regular expressions. 
    The constructed RDF is then integrated into the seed KG following the same steps as in \sspRDFa.
    % \item \sspJSONb~ uses a heuristic for label and property type detection (relation or attribute), it applies text embeddings to directly link JSON objects and keys to existing entities and relations in the given KG. If no label is found, a concatenation of object values is used to generate an entity embedding. Objects (entities) without a sufficiently similar match are assigned new entity identifiers.
    \item \sspJSONc~ avoids explicit mapping generation and instead prompts an LLM to directly output ontology-compliant RDF triples from the input JSON document. The LLM receives both the JSON data and the KG ontology as input and generates triples aligned with the ontology. Unlike approaches that require generating and selecting among candidate mappings, this strategy applies uniformly across all documents, thereby averaging out quality fluctuations in the LLM outputs over the dataset. This removes the need for manual or heuristic mapping selection, though it still introduces variability at the individual document level.
\end{itemize}

\subsection{Unstructured Sources (Text)}

%For unstructured sources, integration is dominated by knowledge extraction and linking, which introduce significant uncertainty and strongly affect overall KG quality. A typical pipeline is:

Unstructured sources such as text and web pages suffer from ambiguity, noise, and limited inherent structure. The main burden of integration falls on natural language processing, which transforms raw text into structured candidate triples. Named entity recognition, relation extraction, and co-reference resolution form the entry point, followed by linking entities and predicates to the target ontology. Knowledge extraction from text is known to be error-prone and highly sensitive to domain, extraction model, and linking strategy, with errors propagating downstream~\cite{DBLP:conf/ccks/CuiLWY17/ReSurvey,regino_systematic_2026,martinez-rodriguez_information_2020,celian_systematic_2025}.
Uncertainty introduced at the extraction stages propagates through the entire pipeline and can limit integration quality. 
%Because of this, knowledge extraction and linking largely determine recall, precision, and runtime. Frequently, intermediate representations in JSON, tabular data, or RDF are produced, enabling reuse of semi-structured or structured integration methods, while inheriting extraction-induced errors.
% KE sentence
%Knowledge extraction from text is known to be error-prone and highly sensitive to domain, extraction model, and linking strategy, with errors propagating downstream~\cite{DBLP:conf/ccks/CuiLWY17/ReSurvey,regino_systematic_2026,martinez-rodriguez_information_2020,celian_systematic_2025}.

A typical pipeline is:

\emph{(i) Knowledge Extraction $\rightarrow$ (ii) Entity/Relation Linking $\rightarrow$ (iii) Schema Alignment $\rightarrow$ (iv) Entity Resolution $\rightarrow$ (v) Cleaning $\rightarrow$ (vi) Completion.}

%\paragraph{TEXT Pipelines}
We use almost similar pipeline layouts for text sources, but with two  different configurations (see~\Cref{fig:ssp_text}):

\begin{itemize}
\item \sspTEXTa~ extracts triple patterns (entity surface forms) from the input text using the tool OpenIE. Entities are linked with DBpedia Spotlight\footnote{DBpedia Spotlight is suitable for entity recognition and linking, because DBpedia widely covers the domain of interest.}, which maps mentions to the seed KG. Mentions not present in the seed are assigned new identifiers in a separate namespace. Relations are then linked using a custom embedding-based relation linker that maps text spans to ontology relations. The  constructed KG is then integrated into the seed KG using the alignment and fusion tasks of \sspRDFa.
% \item \sspTEXTb~ Also applies OpenIE, but for linking, it uses a text embedding model for both relation and entity linking (on labels or all values). Extracted entity mentions are compared with existing KG entities using text embeddings. As this approach links mentions directly to KG entities (ids), it bypasses the need for a separate entity matching or graph alignment step. The resulting triples are integrated into the KG, preserving existing values unless none are available.
\item \sspTEXTc~ replaces OpenIE with an LLM-based triple extraction component, which generates surface-form triples directly from the input text. Entity linking, relation linking, and fusion are performed as in \sspTEXTb.
\end{itemize}

% For all LLM-based tasks, we use OpenAI’s \texttt{gpt-5-mini} model to balance quality, runtime, and cost.

\section{Evaluation}
\label{sec:results}

Our evaluation mainly aims at demonstrating the usability and usefulness of the \bname\   benchmark, by comparatively evaluating the metrics introduced in Section 4 for the datasets and incrementally executed pipelines of the movie domain. In addition of determining the relative effectiveness of different pipeline we also aim at identifying integration problems visible through the produced KG

We focus on integrating three sources (data splits) into the seed KG of \bname-Movie\ by incrementally executing three pipelines to generate a final KG that can be compared to the reference KG. For the three input data formats, we apply the six  single-source type pipelines (SSPs) of Section 6 when all splits have the same format and six multi-source type pipelines (MSP) when each split is of a different format. The MSP pipelines use the base version of the SSP pipelines. For example, the MSP pipelines abbreviated RJT indicates the MSP of the SSP sequence RDFbase$\rightarrow$JSONbase$\rightarrow$TEXTbase.

All pipelines were executed on a machine equipped with an AMD Ryzen 9 9850HX processor (16 cores, 32 threads), 64 GB of main memory, a 1 TB SSD, and an NVIDIA RTX 4070 GPU. The availability of the GPU significantly increased the throughput of embedding-based task implementations. To balance quality and runtime, the LLM-based pipelines use OpenAI’s \texttt{gpt-5-mini-2025-10}, with API costs of RDFllm=0.1€, JSONllm=9.9€, and TEXTllm=2.3€. Prompts are linked in the project repository. 
The LLM-based pipelines for JSON and text had to be executed on the smaller 1k version of the benchmark to limit execution time and monetary expenses. 
All LLM-based tasks use OpenAI’s gpt-5-mini model (API version as of October 2025), with fixed prompts and zero-shot inference.
% \todoi{This is non-negotiable now. Reviewers increasingly expect this, and you already partly do it. "All LLM-based tasks use OpenAI’s gpt-5-mini model (API version as of October 2025), with fixed prompts and zero-shot inference."}
%To limit the number of MSP pipelines, we only consider cases where every split is of a different format and we only consider one of the a and b pipelines per step (RDFa, JSONb, TEXTa).
We apply fixed thresholds\footnote{Thresholds were selected based on preliminary experiments exploring a range of parameter settings} across all pipeline configurations to ensure consistency of evaluation and limit the number of pipeline executions: PARIS matches entities at a similarity threshold of 0.95 and relations at 0.8, 
% JedAI matches at 0.5, Valentine matches at 0.1, 
LLM OM matches at 1.0, DBpedia Spotlight links at 0.8, and an embedding-based relation linker (EmbRL) at 0.8.
% \todoi{eliminate mentioned tools/approaches  not used in the six pipelines}
%Based on early insights and for diversity, we chose RDFa, JSONb, and TEXTa as candidates for the six different source orders of the MSPs.

We next evaluate the integration pipelines using the metrics introduced in Section 4 and also analyze their incremental behavior. Finally, we present comparative rankings of the 12 pipeline combinations. 

\begin{table}
\centering
% \begin{tabular}{lllllll}
\caption{Pipeline result statistics: FC=Fact Count, EC=Entity Count, RC=Relation Count, TC=Type Count, UT=Un-Typed Entities, D=Duration in sec, M=Memory in Gigabyte. * the \sspJSONc~and \sspTEXTc~ pipelines are only executed for the 1k version of the benchmark. }
\begin{tabular}{lrrrrrrr}
\toprule
Pipeline  & FC      & EC     & RC & TC & UT & D(s) & M(GB)  \\
\midrule
Seed 1k & 16,417 & 2,793 & 25 & 3 & - & - & - \\
Seed 10k & 123,686 & 19,527 & 25 & 3 & - & - & - \\
Ref. 1k$^*$   & 63,359  & 8,935  & 25 & 3  & -  & -   & -      \\
Ref. 10k  & 336,002 & 47,706 & 25 & 3  & -   & -  & -     \\
\midrule
\sspRDFa     & 256,528 & 47,258 & 25 & 3  & 3,500    & 65    & 6.3  \\
% R\_B     & 278,366 & 52,932 & 25 & 3  & 7,512    & 6,491 & 44.5 \\
\sspRDFc     & 327,714 & 47,013 & 25 & 3  & 273     & 222   & 6.1  \\
\sspJSONa    & 242,319 & 46,657 & 25 & 3  & 4,801    & 77    & 4.3  \\
% J\_B     & 249,547 & 36,977 & 25 & 3  & 3,359    & 166   & 5.6  \\
\sspJSONc*     & 49,787  & 7,668  & 25 & 3  & 48       & 7,025 & 4.8  \\
\sspTEXTa    & 133,883 & 23,946 & 25 & 3  & 72      & 1,004 & 19.8 \\
% T\_B     & 150,541 & 22,185 & 25 & 3  & 33      & 1,030 & 10.4 \\
\sspTEXTc*     & 24,134  & 3,608  & 25 & 3  & 3       & 1,963 & 10.4 \\
\midrule
RJT$_{base}$     & 228,398 & 37,654 & 25 & 3  & 2,279    & 429   & 19.8 \\
RTJ$_{base}$      & 227,774 & 37,745 & 25 & 3  & 2,308    & 408   & 19.8 \\
JRT$_{base}$      & 227,846 & 38,375 & 25 & 3  & 2,960    & 411   & 19.8 \\
JTR$_{base}$      & 231,706 & 38,541 & 25 & 3  & 2,727    & 398   & 19.8 \\
TJR$_{base}$      & 232,989 & 38,619 & 25 & 3  & 2,655    & 411   & 19.8 \\
TRJ$_{base}$      & 230,945 & 37,980 & 25 & 3  & 2,327    & 417   & 19.8 
% v2 vldb
% J\_A & 242,319 & 46,657 & 25 & 3 & 77    & 4.3  \\
% J\_B & 249,547 & 36,977 & 25 & 3 & 166   & 5.6  \\
% J\_C* & 49,787  & 7,668  & 25 & 3 & 7,025 & 4.8  \\
% R\_A & 256,528 & 47,258 & 25 & 3 & 65    & 6.3  \\
% R\_B & 278,366 & 52,932 & 25 & 3 & 6,491 & 44.5 \\
% R\_C & 327,714 & 47,013 & 25 & 3 & 222   & 6.1  \\
% T\_A & 133,883 & 23,946 & 25 & 3 & 1,004 & 19.8 \\
% T\_B & 150,541 & 22,185 & 25 & 3 & 1,030 & 10.4 \\
% T\_C* & 24,134  & 3,608  & 25 & 3 & 1,963 & 10.4 \\
% \midrule
% JRT  & 227,846 & 38,375 & 25 & 3 & 411   & 19.8 \\
% JTR  & 231,706 & 38,541 & 25 & 3 & 398   & 19.8 \\
% RJT  & 228,398 & 37,654 & 25 & 3 & 429   & 19.8 \\
% RTJ  & 227,774 & 37,745 & 25 & 3 & 408   & 19.8 \\
% TJR  & 232,989 & 38,619 & 25 & 3 & 411   & 19.8 \\
% TRJ  & 230,945 & 37,980 & 25 & 3 & 417   & 19.8 
\\ \bottomrule
\end{tabular}

%The (c)$^2$ was executed 3 times, with the best result selected.} %\todoi{is RDFc also on 1k? Answer: only pipelines marked with $^1$}
% \todoi{Update! values before latest fixes}
\label{tab:overview}
% \todoi{es sollten die gleichen Abkürzungen wie in Kap. 5.2.1 verwendet werden} 
\end{table}

\begin{table*}
\centering
\caption{Quality metrics for all pipelines after the final integration step, combining entity- and fact-level coverage and correctness, with ontology-based consistency scores. The results highlight clear differences between pipeline types, with RDF-based pipelines achieving the highest overall quality, while JSON and especially text-based pipelines show reduced accuracy and coverage due to mapping and extraction errors. Higher values indicate better quality. * the \sspJSONc~and \sspTEXTc~pipelines are only executed for the 1k version of the benchmark.}
\setlength{\tabcolsep}{3pt} % SPECIAL
\begin{tabular}{l|ccc|ccc|c|cccccc}
\toprule
         % & \multicolumn{3}{c|}{Coverage} & \multicolumn{2}{c|}{Correctness} & \multicolumn{7}{c}{Consistency}                    \\
Pipeline & $Cov_E$ & $Corr_E$ & $F1_E$ & $Cov_T$ & $Corr_T$ & $F1_T$ & $1-DR$   & $O_{DT}$ & $O_D$  & $O_R$  & $O_{RD}$ & $O_{LT}$ & $O_{LF}$ \\ % & $Avg_O$  \\
\midrule
\sspRDFa  & 0.858 & 0.875 & 0.866 & 0.541 & 0.901 & 0.676 & 0.995 & 0.994 & 0.995 & 0.988 & 1     & 1     & 1    \\ % & 0.996 \\
\sspRDFc  & 0.959 & 0.988 & 0.973 & 0.884 & 0.917 & 0.901 & 0.995 & 0.993 & 0.996 & 0.989 & 1     & 1     & 1    \\ % & 0.996 \\
\sspJSONa & 0.791 & 0.824 & 0.807 & 0.457 & 0.871 & 0.600 & 0.995 & 0.995 & 0.996 & 0.842 & 1     & 0.794 & 1    \\ % & 0.938 \\
\sspJSONc* & 0.673 & 0.924 & 0.779 & 0.517      &  0.587     & 0.549      & 0.973 & 0.987 & 0.987 & 0.992 & 0.992 & 1     & 1    \\ % & 0.993 \\
\sspTEXTa & 0.020 & 0.285 & 0.037 & 0.009 & 0.191 & 0.017 & 1     & 0.978 & 0.969 & 0.974 & 0.997 & 1     & 0.997 \\ % & 0.986 \\
\sspTEXTc* & 0.042 & 0.322 & 0.074 & 0.044      & 0.160      & 0.069      & 1     & 0.737 & 0.551 & 0.867 & 0.849 & 1     & 0.939 \\ % & 0.824 \\
\midrule
$RJT_{base}$                      & 0.446 & 0.819 & 0.578 & 0.306 & 0.602 & 0.406 & 0.996 & 0.966 & 0.961 & 0.959 & 0.999 & 1     & 1     \\ % & 0.981 \\
$RTJ_{base}$                      & 0.447 & 0.818 & 0.578 & 0.306 & 0.605 & 0.406 & 0.996 & 0.964 & 0.957 & 0.952 & 0.999 & 1     & 0.999 \\ % & 0.979 \\
$JRT_{base}$                      & 0.444 & 0.782 & 0.566 & 0.306 & 0.600 & 0.406 & 0.995 & 0.97  & 0.961 & 0.962 & 0.999 & 1     & 1     \\ % & 0.982 \\
$JTR_{base}$                      & 0.457 & 0.797 & 0.581 & 0.326 & 0.615 & 0.426 & 0.995 & 0.97  & 0.961 & 0.958 & 0.999 & 1     & 0.999 \\ %& 0.981 \\
$TJR_{base}$                      & 0.461 & 0.801 & 0.585 & 0.328 & 0.614 & 0.428 & 0.994 & 0.966 & 0.958 & 0.957 & 0.999 & 1     & 0.999 \\ %& 0.980 \\
$TRJ_{base}$                      & 0.453 & 0.817 & 0.583 & 0.318 & 0.611 & 0.418 & 0.996 & 0.964 & 0.959 & 0.957 & 0.999 & 1     & 0.999 \\ %& 0.980 \\

\bottomrule
\end{tabular}
% \caption{Quality metrics for all pipelines after the final integration step. The table reports entity- and fact-level accuracy (precision, recall, F1) as well as ontology-based consistency scores. Higher values indicate better quality.}
%\todoi{remove 1-DR column?}
\label{tab:quality}
\end{table*}

\subsection{Results}

Table~\ref{tab:overview} summarizes the statistical properties   and Table~\ref{tab:quality} the achieved quality scores of the evaluated pipelines including F1 scores combining coverage and precision values for entities and triples. Note that the \sspJSONc~and \sspTEXTc~ pipelines could only be executed on the smaller 1K version of the benchmark so that comparability of their results is limited. 

The results show clear differences between source formats and between baseline and LLM-enhanced variants.
Among all pipelines, the RDF pipelines achieve the highest overall quality, confirming that structured sources combined with strong matching lead to the most reliable integration. $\mathit{RDF}_{llm}$ achieves the strongest results ($F1_E=0.973$, $F1_T=0.901$), improving over $\mathit{RDF}_{base}$ particularly in fact-level recall. 
\bluenote{
This indicates that improvements in ontology/schema matching propagate to downstream tasks such as entity resolution and fusion, ultimately improving end-to-end KG quality. The task-specific matching results in Appendix (see Table~\ref{tab:matching}) support this observation by showing that RDF$_{llm}$ achieves stronger ontology matching quality while maintaining high precision.
}
% This indicates that improvements in ontology/schema matching propagate to downstream tasks such as entity resolution and fusion, ultimately improving end-to-end KG quality.
% Since the LLM was used for ontology/schema matching this indicates that this task has been much better solved and could thus also improves subsequent tasks like entity resolution and ultimately final KG quality. 
%The near-identical precision values of both variants, together with low duplicate rates, indicate robust fusion and matching behavior.

The JSON pipelines reach a medium quality. $\mathit{JSON}_{base}$ achieves competitive entity coverage, but lower fact recall indicating losses during mapping and relation generation. 
$\mathit{JSON}_{llm}$ is executed on the small benchmark only but is still worse than $\mathit{JSON}_{base}$ especially regarding entity coverage  and fact precision. It also has the highest duplication rate (DR) and by far the highest execution time.

The text pipelines show the weakest performance across both coverage and correctness, with very low entity and triple recall. This directly exposes knowledge extraction and entity linking as the major bottlenecks for unstructured sources. Precision is substantially higher than recall, indicating that extracted assertions are often plausible but highly incomplete. Again the LLM variant $\mathit{JSON}_{llm}$ is very slow despite its execution on the small benchmark.
\bluenote{
In contrast to the RDF pipelines, the LLM-enhanced text pipeline does not consistently improve task-level recall (see Appendix Table~\ref{tab:linking}). Nevertheless, it achieves slightly better end-to-end scores due to producing more precise and semantically consistent extractions, illustrating that improvements in downstream KG quality do not necessarily correlate with higher extraction recall alone.
}

%This distinction would be obscured without evaluating both dimensions jointly.

Differences between entity- and triple-level scores also reveal integration problems. For example, higher entity recall than triple recall indicates that entities may be introduced successfully but remain weakly connected or incompletely populated. Conversely, precision degradation together with duplicate penalties points to entity resolution problems, where unresolved duplicates inflate graph size while reducing semantic quality.

%\paragraph{Consistency metrics expose structural integration defects.}
The ontology-based consistency metrics provide complementary insights not captured by coverage or correctness alone. The RDF pipelines show near-perfect consistency, indicating that structured integration largely preserves ontology constraints. Hence, their strong coverage and correctness are not achieved at the expense of structural correctness.
In contrast, JSON and especially text pipelines show lower compliance in domain/range constraints and relation usage. These violations indicate concrete integration defects, such as incorrectly mapped predicates, missing or wrong types, or semantically invalid links. 
$\mathit{JSON}_{base}$ has most violations for relation direction $\mathit{O}_{R}$ and literal datatypes $\mathit{O}_{LT}$ as well as the highest number of un-typed entities (Table~\ref{tab:overview}) indicating problems in data mapping to RDF and type inference. 
For the text pipelines, consistency degradations align with the weak extraction and linking performance seen in the coverage and correctness metrics, showing how early-stage extraction errors propagate into KG consistency problems.

%This demonstrates an important property of the benchmark: consistency does not merely provide an additional score, but helps explain \emph{why} pipelines with moderate coverage or precision may still produce lower-quality KGs.

The six MSP pipelines perform  similarly and reach  modest
F1 scores especially since there is always a low-quality text pipeline involved.  Their man problem is a low coverage while precision values are competitively 
high, especially for triples.

The  order in which 
the source formats are integrated, e.g., whether text data is integrated first or last,
seems to have little influence on the final KG quality. 

\paragraph{Incremental behavior.}
% \begin{figure}[!ht]
%     \centering
%     \includegraphics[width=1\linewidth]{figures/test_fig_both_growth.png}
%     \caption{Visualization of statistical metrics (growth) for the 12 pipelines and their three increments/stages $KG_1-KG_3$. The black dotted lines indicate the expected reference KG sizes. All three SSP (c) pipelines are omitted here.}
%     \label{fig:growth}
% \end{figure}
\begin{figure*} 
    \centering
    \includegraphics[width=1\linewidth]{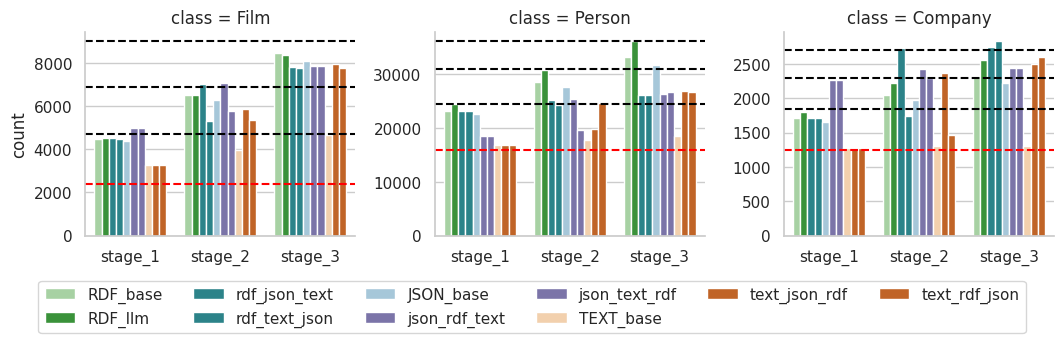}
    \caption{Number of integrated entities by entity type and expected entities at each source increment (stage) for all pipelines. The dotted lines indicate the values of the current reference between the seed (red) and the three source increments (black).} %\todoi{films should come first, companies last}
    \label{fig:entity_classes}
% \todoi{start with RDF pipelines, followed by JSON and Text pipelines. Leave out small cases JSONllm and TEXTllm}
\end{figure*}

Figure~\ref{fig:entity_classes} shows how the size of the integrated KG  evolves over successive source integrations for the three classes Film, Person and Company. RDF pipelines maintain stable increases across stages and can 
nearly match the size of the reference graphs indicating a continuous high coverage. 
By contrast, the pure text pipelines can hardly add further
entities  compared to the seed graph for Person and company entities.
The MSP pipelines  struggle the most for entities of the most frequent class \texttt{Person}   since both JSON and TEXT steps have difficulties with this entity type. 

\paragraph{Summary of findings.}
Overall, the evaluation shows integrating structured (RDF) data has clear advantages over the integration of semi-structured and text data. LLM  use  was very beneficial for RDF input but very slow and ineffective for JSON and text data. 
The observed limitations also show that most 
pipelines and their configurations are far from perfect providing room for improvement.  
The proposed benchmark  metrics proved to be useful to identify integration problems and pipeline weaknesses. Low coverage exposes incomplete extraction or mapping, reduced precision and duplicate penalties expose linking and fusion errors, and consistency violations reveal ontology-level defects. Across all pipelines, entity resolution and schema alignment emerge as major drivers of KG quality, while knowledge extraction remains the principal bottleneck for unstructured data. 
%These findings illustrate how the benchmark supports not only comparative pipeline evaluation, but also systematic diagnosis of the causes of quality differences.

\bluenote{
\subsection{Comparative rankings}
}
\label{sec:ranking}

To compare pipelines under different evaluation priorities, we aggregate the three main quality dimensions introduced in~\Cref{sec:eval_metrics}, namely coverage, correctness, and consistency. For each dimension, we first compute the arithmetic mean of the corresponding subgroup metrics: entity and triple recall for coverage, entity and triple precision for correctness, and the ontology consistency metrics for consistency. We then enumerate all weight combinations for the three quality scores using a step size of 0.1, resulting in 66 distinct weight permutations. \Cref{tab:ranking} summarizes the resulting aggregated subgroup scores together with their harmonic mean and the observed rank ranges across all weight permutations.

The results reveal a clear hierarchy of pipeline performance. The \textit{RDF}$_{llm}$ pipeline consistently achieves the best results across all quality dimensions, resulting in the highest harmonic mean (0.956) and a stable rank range of only 1--1. This indicates that the pipeline remains the top-ranked approach independently of the weighting configuration.

The  \textit{RDF}$_{base}$ pipeline also performs strongly, achieving the second-highest harmonic mean (0.843) with a fixed rank range of 2--2, confirming the robustness of RDF-based integration approaches.

The JSON pipelines achieve intermediate results. While \textit{JSON}$_{base}$ obtains a slightly higher harmonic mean than \textit{JSON}$_{llm}$, its much larger rank range (3--11) indicates a stronger dependence on the weighting configuration. In contrast, \textit{JSON}$_{llm}$ exhibits more stable rankings despite slightly lower aggregated quality values.

The MSP pipelines occupy the middle range of the ranking with harmonic means between approximately 0.585 and 0.604. Their relatively narrow score differences indicate that the integration order has only limited influence on the final KG quality, although moderate rank variability remains depending on the weighting priorities.

The text-based pipelines achieve the lowest overall results. In particular, \textit{TEXT}$_{base}$ and \textit{TEXT}$_{llm}$ obtain very low harmonic means due to their limited coverage and correctness despite comparatively high consistency values. This shows that knowledge extraction and linking remain the primary bottlenecks for unstructured source integration.

% Overall, the ranking results demonstrate that pipelines operating directly on structured RDF sources provide the most robust and consistently high-quality integration results across varying evaluation priorities.

Beyond comparing pipelines, the ranking analysis also illustrates an important property of the benchmark: pipeline comparisons remain largely stable under changing user priorities. This indicates that the benchmark does not overfit to a particular weighting choice and that observed quality differences reflect robust pipeline characteristics rather than artifacts of score aggregation.

Taken together, the comparative rankings reinforce the conclusions from the metric-level analysis: structured integration pipelines provide the most reliable KG quality, LLM enhancements can improve specific integration stages depending on the source type, and weaknesses in extraction or matching systematically propagate into lower end-to-end rankings.

\begin{table}
\centering
\caption{Comparative ranking of pipelines based on the aggregated subgroup scores for coverage, accuracy, and consistency. The Harmonic Mean summarizes the three quality dimensions while penalizing imbalanced performance. The Rank Range column shows the observed rank interval across all weight permutations used for the total score calculation, indicating the robustness of pipeline rankings under different evaluation priorities.}
\begin{tabular}{l|cccc|c}
\toprule
Pipeline  & Cov. & Corr. & Cons. & H-Mean & Rank Range  \\
\midrule
\sspRDFc  & 0.921    & 0.953    & 0.996       & 0.956 & 1 - 1        \\
\sspRDFa  & 0.7      & 0.888    & 0.996       & 0.843 & 2 - 2        \\
\sspJSONa & 0.624    & 0.847    & 0.946       & 0.781 & 3 - 11       \\
\sspJSONc* & 0.595    & 0.756    & 0.99        & 0.747 & 3 - 4        \\
$TJR_{base}$       & 0.395    & 0.708    & 0.982       & 0.604 & 5 - 9        \\
$JTR_{base}$       & 0.392    & 0.706    & 0.983       & 0.601 & 5 - 8        \\
$TRJ_{base}$       & 0.386    & 0.714    & 0.982       & 0.598 & 4 - 9        \\
$RTJ_{base}$       & 0.377    & 0.712    & 0.981       & 0.59  & 4 - 9        \\
$RJT_{base}$       & 0.376    & 0.71     & 0.983       & 0.59  & 6 - 10       \\
$JRT_{base}$       & 0.375    & 0.691    & 0.984       & 0.585 & 5 - 10       \\
\sspTEXTc* & 0.043    & 0.241    & 0.849       & 0.105 & 4 - 12       \\
\sspTEXTa & 0.014    & 0.238    & 0.988       & 0.04  & 11 - 12      \\
\bottomrule
\end{tabular}
\label{tab:ranking}
\end{table}

\section{Conclusions and Outlook}
\label{sec:conclusion}

We presented \bname, a benchmark for the end-to-end evaluation of knowledge graph integration pipelines. In contrast to prior work that primarily evaluates isolated tasks such as extraction, matching, or ontology alignment, our approach focuses on assessing the quality of the integrated knowledge graph produced by complete pipelines or sequences of several pipelines. \bname\ is based on the three complementary metrics coverage, correctness, and consistency to comprehensively evaluate the quality of the integrated KG. Furthermore,  
we provide the open domain-specific benchmark  \bname-Movie\   for incremental integration of heterogeneous sources and comparisons with a reference KG. 

We demonstrated the usefulness of the new benchmarks by a comparative  evaluation of baseline and LLM-enhanced pipelines for RDF, JSON, and text sources. The results demonstrate clear differences in integration quality across source types and pipeline designs. Structured RDF pipelines achieved the strongest and most robust results, while semi-structured and especially text-based pipelines revealed substantial challenges in mapping, extraction, and linking. The evaluation further showed that improvements in individual tasks, such as schema matching, can measurably improve end-to-end KG quality, while weaknesses in a single stage can dominate overall pipeline performance. These findings underline the importance of evaluating integration pipelines holistically rather than through isolated task metrics alone.
The evaluation also showed that the proposed metrics can serve as diagnostic indicators of characteristic integration problems. Coverage metrics expose incompleteness, correctness metrics reveal matching and fusion errors, and consistency metrics capture structural and semantic defects that would otherwise remain hidden.

There are several opportunities for future work. There are many ways to improve pipeline design,  including additional LLM-based approaches and more advanced fusion or reasoning techniques, and test these with the existing benchmark. The benchmark can be extended beyond the movie domain to additional domains and ontologies in order to study additional integration challenges and increase  the generality of the evaluation methodology.  
It is furthermore promising to investigate automated support for diagnosing pipeline weaknesses and guiding the improvement of pipeline designs based on the observed evaluation patterns.

\begin{acks}
The authors acknowledge the financial support by the Federal Ministry of Education and Research of Germany and by the Sächsische Staatsministerium für Wissenschaft Kultur und Tourismus in the program Center of Excellence for AI-research "Center for Scalable Data Analytics and Artificial Intelligence Dresden/Leipzig", project identification number: ScaDS.AI.
\end{acks}

\bibliographystyle{ACM-Reference-Format}
\bibliography{research}

\newpage
\section*{Appendix}

\Cref{tab:matching,tab:linking} report task-specific evaluation results for the matching and linking tasks used in the evaluated pipelines.
The task-specific results indicate that $\mathit{RDF}_{llm}$ achieves substantially higher ontology matching recall than $\mathit{RDF}_{base}$ while maintaining high precision. This improves downstream entity matching and fusion because more semantically corresponding relations become available during graph alignment. As a consequence, the pipeline integrates more correct triples and achieves higher end-to-end KG coverage and F1 scores.

\begin{table}[h!]
    \centering
    \caption{Entity matching and ontology matching scores (precision and recall) across all entity types.}
\begin{tabular}{llGGGG}
\toprule
Pipeline    & inc. & \multicolumn{1}{c}{$EM_{p}$} & \multicolumn{1}{c}{$EM_{r}$} & 
\multicolumn{1}{c}{$OM_{p}$} &  \multicolumn{1}{c}{$OM_{r}$} \\
\midrule
\sspRDFa & 1 & 0.98 & 1.00 & 1.00 & 0.63 \\
\sspRDFa & 2 & 0.98 & 1.00 & 1.00 & 0.58 \\
\sspRDFa & 3 & 0.98 & 1.00 & 1.00 & 0.63 \\
% R\_B & 1 & 0.83 & 0.88 & 0.89 & 0.73 \\
% R\_B & 2 & 0.81 & 0.88 & 0.89 & 0.77 \\
% R\_B & 3 & 0.82 & 0.88 & 0.89 & 0.77 \\
\sspRDFc & 1 & 0.98 & 1.00 & 1.00 & 0.83 \\
\sspRDFc & 2 & 0.98 & 1.00 & 1.00 & 0.83 \\
\sspRDFc & 3 & 0.98 & 1.00 & 1.00 & 0.75 \\
\sspJSONa & 1 & 0.99 & 0.41 & 0.43 & 0.38 \\
\sspJSONa & 2 & 0.75 & 0.40 & 0.43 & 0.38 \\
\sspJSONa & 3 & 0.66 & 0.40 & 0.43 & 0.38 \\
\midrule
$R\_\__{base}$  & 1 & 0.98 & 1.00 & 1.00 & 0.63 \\
$JRT_{base}$  & 2 & 0.97 & 1.00 & 1.00 & 0.58 \\
$JTR_{base}$  & 3 & 0.97 & 1.00 & 1.00 & 0.58 \\
$TJR_{base}$  & 3 & 0.97 & 1.00 & 1.00 & 0.63 \\
$TRJ_{base}$  & 2 & 0.98 & 1.00 & 1.00 & 0.58 
\\ \bottomrule
\end{tabular}
\label{tab:matching}
\end{table}

\begin{table}[h!]
    \centering
    \caption{Entity linking recall scores for film entities extracted from text documents. Evaluation is restricted to film entities; person and company entities yield even lower results.}
\begin{tabular}{llG}
\toprule
Pipeline  & inc. & \multicolumn{1}{c}{$EL_r$}    \\
\midrule
% J\_B & 1 & 0.64 \\
% J\_B & 2 & 0.62 \\
% J\_B & 3 & 0.62 \\
\sspTEXTa & 1 & 0.43 \\
\sspTEXTa & 2 & 0.43 \\
\sspTEXTa & 3 & 0.45 \\
% T\_B & 1 & 0.13 \\
% T\_B & 2 & 0.12 \\
% T\_B & 3 & 0.12 \\
\sspTEXTc* & 1 & 0.21 \\
\sspTEXTc* & 2 & 0.14 \\
\sspTEXTc* & 3 & 0.16 \\
\midrule
$RTJ_{base}$  & 2 & 0.43 \\
$RJT_{base}$  & 3 & 0.45 \\
$JTR_{base}$  & 2 & 0.43 \\
$JRT_{base}$  & 3 & 0.45 \\
$T\_\__{base}$  & 1 & 0.43 
\\ \bottomrule
\end{tabular}
\label{tab:linking}
\end{table}

\end{document}